\PassOptionsToPackage{table}{xcolor}
\documentclass[11pt]{article}

\usepackage[preprint]{acl}

\usepackage{times}
\usepackage{latexsym}

\usepackage[T1]{fontenc}
\usepackage[utf8]{inputenc}

\usepackage{microtype}

\usepackage{inconsolata}

\usepackage{graphicx}

\usepackage{amsmath}
\usepackage{amssymb}
\usepackage{mathtools}
\usepackage{tikz}
\usetikzlibrary{
  positioning,
  patterns,
  decorations.pathreplacing,
  calc
}
\usepackage{booktabs}
\usepackage{siunitx}
\usepackage{multirow}
\usepackage{adjustbox}
\usepackage{makecell}
\usepackage{algorithm}
\usepackage{algpseudocode}
\usepackage[most]{tcolorbox}
\tcbuselibrary{listings}
\usepackage{xspace}
\usepackage{adjustbox}
\usepackage{subcaption}
\usepackage{enumitem}
\usepackage{rotating}

\usepackage[textsize=footnotesize]{todonotes}

\DeclareMathOperator*{\argmax}{arg\,max}

\definecolor{orangeSpan}{RGB}{245, 166, 35}
\definecolor{greenSpan}{RGB}{80, 180, 80}
\definecolor{highlightT}{RGB}{230, 210, 170}
\definecolor{highlightS}{RGB}{170, 210, 170}

\newcommand{\tp}{\text{tp}}

\newcommand{\fp}{\text{fp}}
\newcommand{\fn}{\text{fn}}

\newcommand{\exm}{\textsc{em}\xspace}
\newcommand{\mpa}{\textsc{mp}\xspace}
\newcommand{\mpp}{\textsc{mpp}\xspace}

\newcommand{\fscore}{F\text{-score}\xspace}

\newcommand{\hyp}{\hat{s}}
\newcommand{\hyps}{\hat{S}}

\newcommand{\langpair}[2]{\textsc{#1}$\rightarrow$\textsc{#2}}
\newcommand{\bs}[1]{\boldsymbol{#1}}

\definecolor{SeaGreen3}{RGB}{67, 205, 128}
\definecolor{Firebrick3}{RGB}{205, 38, 38}
\definecolor{highlight}{gray}{0.90}

\newcommand{\rankcolor}[1]{%
  \ifcase#1\or SeaGreen3!100!Firebrick3!50\or SeaGreen3!92!Firebrick3!50\or SeaGreen3!83!Firebrick3!50\or SeaGreen3!75!Firebrick3!50\or SeaGreen3!67!Firebrick3!50\or SeaGreen3!58!Firebrick3!50\or SeaGreen3!50!Firebrick3!50\or SeaGreen3!42!Firebrick3!50\or SeaGreen3!33!Firebrick3!50\or SeaGreen3!25!Firebrick3!50\or SeaGreen3!17!Firebrick3!50\or SeaGreen3!8!Firebrick3!50\or SeaGreen3!0!Firebrick3!50\fi
}
\newcommand{\rc}[1]{\cellcolor{\rankcolor{#1}}#1}

\newcommand{\minor}{\texttt{minor}\xspace}
\newcommand{\major}{\texttt{major}\xspace}

\newcommand{\matese}{\text{MaTESe}\xspace}

\newcommand{\xcometxl}{\text{XCOMET-XL}\xspace}

\newcommand{\gptoss}{\text{gpt-oss 120b}\xspace}
\newcommand{\sonnet}{\text{Claude Sonnet 4.5}\xspace}
\newcommand{\haiku}{\text{Claude Haiku 4.5}\xspace}
\newcommand{\qwen}{\text{Qwen3 235b}\xspace}

\title{Span-Level Machine Translation Meta-Evaluation}

\author{Stefano Perrella\textsuperscript{*} \\ Sapienza University of Rome \\ \texttt{perrella@diag.uniroma1.it} \\\And
  Eric Morales Agostinho \\ Amazon \\ \texttt{ericmrls@amazon.com} \\\And Hugo Zaragoza \\ Amazon \\ \texttt{hugzarag@amazon.com}}

\begin{document}
\maketitle

\def\thefootnote{*}\footnotetext{Work conducted while the author was at Amazon.}
\def\thefootnote{\arabic{footnote}}

\begin{abstract}
Machine Translation (MT) and automatic MT evaluation have improved dramatically in recent years, enabling numerous novel applications. Automatic evaluation techniques have evolved from producing scalar quality scores to precisely locating translation errors and assigning them error categories and severity levels. However, it remains unclear how to reliably measure the evaluation capabilities of auto-evaluators that do error detection, as no established technique exists in the literature. This work investigates different implementations of span-level precision, recall, and $\fscore$, showing that seemingly similar approaches can yield substantially different rankings, and that certain widely-used techniques are unsuitable for evaluating MT error detection. We propose ``match with partial overlap and partial credit'' (\mpp) with micro-averaging as a robust meta-evaluation strategy and release code for its use publicly. Finally, we use \mpp to assess the state of the art in MT error detection.
\end{abstract}

    % * P0: remove things/appendix
    % * P0: include sections comparing with humans.
    %     * share data about this
    % * P0: what happens when you don’t match with 1:1 matching.
    %     * no data, only arguments.
    % * P1: iterative experiment when you keep increasing the spans detected by the extended.
    %     * it is implemented but it is not run already.
    % * P2: normal evaluator, create the low recall, but remove errors randomly to show that the problem is that there are samples with no errors.
    %     * not implemented and no data (it was implemented partially at some point, but we need to remove the same number of errors we are removing in the base low recall).

\begin{table}[t]
\centering
\resizebox{0.90\columnwidth}{!}{
\begin{tabular}{l|rrrrrrrr}
\toprule
& \multicolumn{4}{c}{Micro} & \multicolumn{4}{c}{Macro} \\
\textbf{Auto-evaluator} & \rotatebox{90}{\exm} & \rotatebox{90}{\mpa} & \rotatebox{90}{\mpp} & \rotatebox{90}{\textsc{w}25} & \rotatebox{90}{\exm} & \rotatebox{90}{\mpa} & \rotatebox{90}{\mpp} & \rotatebox{90}{\textsc{w}25} \\
\midrule
\cellcolor{highlight}\qwen & \rc{5} & \rc{1} & \rc{1} & \rc{1} & \rc{12} & \rc{12} & \rc{10} & \rc{12} \\
GemSpanEval.sec & \rc{3} & \rc{2} & \rc{2} & \rc{2} & \rc{4} & \rc{4} & \rc{4} & \rc{4} \\
\cellcolor{highlight}Sonnet 4.5 & \rc{2} & \rc{5} & \rc{3} & \rc{6} & \rc{8} & \rc{11} & \rc{7} & \rc{9} \\
GemSpanEval.pri & \rc{4} & \rc{4} & \rc{4} & \rc{5} & \rc{6} & \rc{5} & \rc{5} & \rc{5} \\
XCOMET-XXL & \rc{11} & \rc{3} & \rc{5} & \rc{4} & \rc{11} & \rc{7} & \rc{11} & \rc{7} \\
AIP.sec & \rc{1} & \rc{7} & \rc{6} & \rc{12} & \rc{3} & \rc{3} & \rc{3} & \rc{3} \\
\cellcolor{highlight}Haiku 4.5 & \rc{7} & \rc{9} & \rc{7} & \rc{7} & \rc{5} & \rc{6} & \rc{6} & \rc{6} \\
XCOMET-XL & \rc{12} & \rc{6} & \rc{8} & \rc{3} & \rc{13} & \rc{10} & \rc{13} & \rc{10} \\
\cellcolor{highlight}gpt-oss 120b & \rc{6} & \rc{10} & \rc{9} & \rc{8} & \rc{10} & \rc{13} & \rc{12} & \rc{13} \\
AIP.pri & \rc{8} & \rc{8} & \rc{10} & \rc{13} & \rc{2} & \rc{2} & \rc{2} & \rc{2} \\
AutoLQA.pri & \rc{9} & \rc{11} & \rc{11} & \rc{9} & \rc{9} & \rc{9} & \rc{9} & \rc{11} \\
AutoLQAESA.sec & \rc{10} & \rc{12} & \rc{12} & \rc{10} & \rc{7} & \rc{8} & \rc{8} & \rc{8} \\
AutoLQA41.sec & \rc{13} & \rc{13} & \rc{13} & \rc{11} & \rc{1} & \rc{1} & \rc{1} & \rc{1} \\
\bottomrule
\end{tabular}
}
\caption{Ranking of auto-evaluators on the MQM split of the WMT 2025 Automated Translation Quality Evaluation Systems Shared Task under the measures defined in Section~\ref{sec:span-level-meta-evaluation} (\exm, \mpa, and \mpp), alongside \textsc{w}25, that is, the measure used at WMT25 (illustrated in Section~\ref{apx:wmt25_measure}), both micro- and macro-averaging results. We run the auto-evaluators highlighted in gray, the others are submissions to WMT.}
\label{tab:wmt25-mqm-ranks}
\end{table}

\section{Introduction}
Machine Translation (MT) evaluation involves assessing the quality of translated text, and automatic evaluation techniques (hereafter, auto-evaluators\footnote{Automatic MT evaluation techniques are often referred to as MT metrics. However, to prevent confusion with the precision, recall, and $\fscore$ metrics, we refer to automatic evaluation techniques exclusively as auto-evaluators.}) assess translation quality without human intervention. Automatic evaluation enables faster and cheaper experimentation when developing translation models compared to human evaluation; also, high-accuracy auto-evaluators are used for various downstream applications such as data filtering, translation re-ranking, and translation difficulty estimation \cite{10.1162/tacl_a_00491, fernandes-etal-2022-quality, perrella-etal-2024-beyond, jon-etal-2025-cuni, kocmi-etal-2025-command, finkelstein-etal-2025-google, tan-2025-simple, garcia-gilabert-etal-2025-salamandra, proietti-etal-2025-estimating}, or as a proxy to fine-tune MT models using Reinforcement Learning objectives \cite{he-etal-2024-improving, 10.5555/3692070.3694345, jon-etal-2025-cuni, zheng-etal-2025-shy, finkelstein-etal-2025-google}. 

Early auto-evaluators assessed translation quality using heuristics based on word n-grams and character-based overlap between a translation and one or more manually curated references \cite{papineni-etal-2002-bleu, banerjee-lavie-2005-meteor, popovic-2015-chrf}. Later, neural auto-evaluators enabled assessing translation quality at a deeper semantic level \cite{rei-etal-2020-comet, wan-etal-2022-unite, rei-etal-2022-cometkiwi, juraska-etal-2023-metricx, proietti-etal-2026-pear}. However, most auto-evaluators return their evaluation as a scalar quality score, which can be difficult to interpret \cite{perrella-etal-2024-beyond}. To mitigate this issue, recent auto-evaluators have been developed to locate translation errors, optionally also classifying them based on category and severity \cite{perrella-etal-2022-matese, kocmi-federmann-2023-gemba, guerreiro-etal-2024-xcomet, juraska-etal-2025-metricx, yeom-etal-2025-tagged, hrabal-etal-2025-cuni}. However, due to the lack of established span-level meta-evaluation strategies -- that is, techniques for evaluating the ability of auto-evaluators to detect translation errors -- prior work has adopted disparate approaches, making results difficult to compare across studies. Furthermore, no previous work has systematically examined the effectiveness and fairness of these approaches, leaving it unclear whether they are equivalent or, most importantly, whether they can be reliably applied to MT error detection.\footnote{We provide an overview of the approaches used in previous works in Section~\ref{sec:previous-work}.}

This work examines different implementations of span-level precision, recall, and $\fscore$, demonstrating that apparently similar approaches conceal arbitrary methodological choices that can lead to drastic differences in the results (Table~\ref{tab:wmt25-mqm-ranks}). Moreover, we find that some commonly employed techniques are unsuitable for evaluating error detection capabilities, resulting in confounded results. We identify ``match with partial overlap and partial credit'' (\mpp), paired with micro-averaging results across data samples, as a robust strategy for span-level MT meta-evaluation, and use it to assess the state of the art in MT error detection. We release code for reproducing our results at \url{https://github.com/amazon-science/span-mt-metaeval}, as well as code for using \mpp at \url{https://github.com/sted19/span-mt-metrics-eval}. 

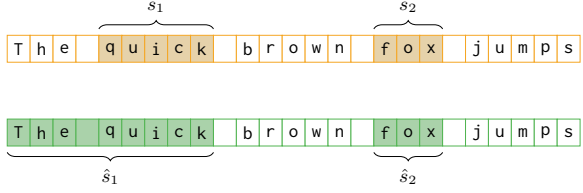
\begin{figure}
    \centering
    \resizebox{\linewidth}{!}{
    \begin{tikzpicture}[
        cell/.style={
          minimum width=0.38cm,
          minimum height=0.45cm,
          draw=#1,
          line width=0.3pt,
          outer sep=0pt,
          inner sep=0pt,
          font=\ttfamily\small,
          anchor=center
        },
        T empty/.style={cell=orangeSpan, fill=white},
        T filled/.style={cell=orangeSpan, fill=highlightT},
        S empty/.style={cell=greenSpan, fill=white},
        S filled/.style={cell=greenSpan, fill=highlightS},
        brace above/.style={
          decorate,
          decoration={brace, amplitude=4pt, raise=3pt}
        },
        brace below/.style={
          decorate,
          decoration={brace, amplitude=4pt, raise=3pt, mirror}
        },
        span label/.style={font=\footnotesize}
      ]
    
      \def\cellwidth{0.38}
      \def\rowsep{1.4}
      
      % =============================================
      % Character data: char/T-highlight/S-highlight
      % =============================================
      
      \foreach \char/\tstat/\sstat [count=\i from 0] in {
        T/0/1, h/0/1, e/0/1, { }/0/1,           % 0-3:   "The "
        q/1/1, u/1/1, i/1/1, c/1/1, k/1/1,      % 4-8:   "quick"
        { }/0/0,                                 % 9:     " "
        b/0/0, r/0/0, o/0/0, w/0/0, n/0/0,      % 10-14: "brown"
        { }/0/0,                                 % 15:    " "
        f/1/1, o/1/1, x/1/1,                    % 16-18: "fox"
        { }/0/0,                                 % 19:    " "
        j/0/0, u/0/0, m/0/0, p/0/0, s/0/0       % 20-24: "jumps"
        % { }/0/0,                                 % 25:    " "
        % o/0/0, v/0/0, e/0/0, r/0/0,             % 26-29: "over"
        % { }/0/0,                                 % 30:    " "
        % t/0/0, h/0/0, e/0/0,                    % 31-33: "the"
        % { }/0/0,                                 % 34:    " "
        % l/1/0, a/1/0, z/1/0, y/1/0,             % 35-38: "lazy"
        % { }/0/0,                                 % 39:    " "
        % d/0/1, o/0/1, g/0/1                     % 40-42: "dog"
      } {
        % T row (top)
        \ifnum\tstat=1
          \node[T filled] (T\i) at (\i*\cellwidth, 0) {\char};
        \else
          \node[T empty] (T\i) at (\i*\cellwidth, 0) {\char};
        \fi
        % S row (bottom)
        \ifnum\sstat=1
          \node[S filled] (S\i) at (\i*\cellwidth, -\rowsep) {\char};
        \else
          \node[S empty] (S\i) at (\i*\cellwidth, -\rowsep) {\char};
        \fi
      }
    
      % =============================================
      % T ANNOTATIONS (top braces)
      % Draw left to right so brace bulges upward
      % =============================================
      \draw[brace above] (T4.north west) -- (T8.north east)
        node[midway, above=7pt, span label] {$s_1$};
      \draw[brace above] (T16.north west) -- (T18.north east)
        node[midway, above=7pt, span label] {$s_2$};
    
      % =============================================
      % S ANNOTATIONS (bottom braces)
      % Using mirror decoration so brace bulges downward
      % =============================================
      \draw[brace below] (S0.south west) -- (S8.south east)
        node[midway, below=7pt, span label] {$\hat{s}_1$};
      \draw[brace below] (S16.south west) -- (S18.south east)
        node[midway, below=7pt, span label] {$\hat{s}_2$};
    
    \end{tikzpicture}
    }
    \caption{Span-level annotations for the translation $\bs{x}=$ ``The quick brown fox jumps''. $\hyp_1, \hyp_2 \in \hyps$ are hypothesis error spans, while  $s_1, s_2 \in S$ are the ground-truth ones.}
    \label{fig:example1}
\end{figure}

\section{Span-level MT Meta-Evaluation} \label{sec:span-level-meta-evaluation}
We formulate MT error detection as a structured prediction task in which auto-evaluators and human annotators identify sequences of characters corresponding to translation errors. Accordingly, span-level MT meta-evaluation aims to quantify the agreement between auto-evaluators and humans by estimating the similarity between their respective error sequences (or error spans). However, methodological choices regarding how this similarity is defined and computed can give rise to metrics that are fundamentally different or, in some cases, even unsuitable for MT error detection. This work focuses on implementations of precision, recall, and $\fscore$, as these metrics are widely used to evaluate MT error detection performance. Nevertheless, most of our observations are broadly applicable and extend to other evaluation measures as well.  

Let $\boldsymbol{x} = (x_1, x_2, \dots, x_n)$ denote a translation consisting of $n$ characters.  A span $s = (i_s, j_s)$ denotes the character subsequence $(x_{i_s}, x_{i_s+1}, \dots, x_{j_s})$, and we denote with $[\![ s ]\!] = \{i_s, i_s+1, \ldots, j_s\}$ the set of character positions covered by $s$. We define the length of $s$ as $\left| s \right| \coloneqq \left| [\![ s ]\!] \right|$. For two spans $s_1$ and $s_2$, we define their intersection and union lengths as:
\begin{equation*}
\resizebox{0.95\linewidth}{!}{$
| s_1 \cap s_2 | \coloneqq |[\![s_1]\!] \cap [\![s_2]\!]|,\quad
| s_1 \cup s_2 | \coloneqq |[\![s_1]\!] \cup [\![s_2]\!]|
$}
\end{equation*}
Let $\mathcal{I}_n = \{ (i,j)\in \mathbb{N}^2 : 1 \leq i \leq j \leq n \}$ denote the set of all valid spans in $\bs{x}$, and let $\hyps \subseteq \mathcal{I}_n$ denote the set of hypothesis error spans -- e.g., the spans detected by an auto-evaluator -- while $S \subseteq \mathcal{I}_n$ is the set of ground-truth error spans.

We wish to compute the precision, recall, and $\fscore$ of $\hyps$ with respect to $S$. However, a hypothesis span may overlap with multiple ground-truth spans, and vice versa, so the correspondence between hypothesis and ground-truth error spans is not unique. We resolve this ambiguity by finding the one-to-one matching between hypothesis and ground-truth error spans that maximizes $\fscore$.\footnote{We refer the reader to Appendix~\ref{apx:enforcing-one-to-one} for a discussion about the effects of enforcing a one-to-one matching.} This way, we also avoid underestimating $\fscore$ due to an unfavorable alignment between hypothesis and ground-truth spans.
Formally, we define the set of all one-to-one matchings between $\hyps$ and $S$:
\begin{equation*}
\mathcal M
=
\left\{
M\subseteq \hyps \times S \ \middle|\
\substack{
\forall \hyp \in \hyps:\ |\{s : (\hyp,s)\in M\}| \le 1\\
\forall s \in S:\ |\{\hyp : (\hyp,s)\in M\}| \le 1
}
\right\}
\end{equation*}
We then select $M^* \in \mathcal{M}$ as the matching that maximizes $\fscore$.

In the following sections, we define several variants of precision, recall, and $\fscore$, which differ in what qualifies as a valid match and whether matches receive binary credit (either full or zero) or may instead be assigned partial credit.

\subsection{Exact Match (\exm)} \label{sec:exact_match}
Under exact match (\exm), $\hyp$ matches $s$ only if the two spans have the same start and end character indices; equivalently, if $\hyp = s$.
We define the set of one-to-one exact-match matchings as follows:
\begin{equation*}
    \mathcal{M}_{\exm} = \left\{ M \in \mathcal{M} \ \middle|\ 
    \forall (\hyp, s) \in M: \hyp = s 
    \right\}
\end{equation*}
Then, given any $M \in \mathcal{M}_{\exm}$, we define precision, recall, and $\fscore$ as follows:
\begin{align} 
    \label{eq:em_precision}
    P_{\exm} &= \begin{cases}
        \frac{\tp}{\tp + \fp} = \frac{|M|}{|\hyps|}, & \text{if } \hyps \neq \emptyset \\
        1 & \text{otherwise}.
    \end{cases} \\ 
    \label{eq:em_recall}
    R_{\exm} &= \begin{cases}
        \frac{\tp}{\tp + \fn} = \frac{|M|}{|S|}, & \text{if } S \neq \emptyset \\
        1 & \text{otherwise}.
    \end{cases} \\ 
    \label{eq:em_fscore}
    F_{\exm} &= \begin{cases}
        \frac{2P_{\exm}R_{\exm}}{P_{\exm} + R_{\exm}}, & \text{if } P_{\exm} + R_{\exm} > 0 \\
        0 & \text{otherwise.}
    \end{cases} 
\end{align}
By setting $P=1$ when $\hyps=\emptyset$ and $R=1$ when $S=\emptyset$, we ensure that an auto-evaluator receives $F=1$ when it detects zero errors in a translation containing no ground-truth errors, while receiving $F=0$ when only one of $\hyps$ and $S$ is empty. Finally, we select $M_{\exm}^* \in \mathcal{M}_{\exm}$ as the one that maximizes $F_{\exm}$.

Let us consider the example in Figure~\ref{fig:example1}. $|\hyps|=|S|=2$, since there are two hypothesis spans and two ground-truth spans. Moreover, $\mathcal{M}_{\exm} = \{ \emptyset, \{ (\hyp_2,s_2) \} \}$, because there are two exact-match one-to-one error matchings. For $M_1=\emptyset$, we have $P_{\exm}=R_{\exm}=F_{\exm} = 0$, while for $M_2=\{ (\hyp_2,s_2) \}$, $P_{\exm} = R_{\exm} = F_{\exm} = \frac{1}{2}$. Thus, we select $M^*_{\exm} = M_2$ as the matching that maximizes $F_{\exm}$.

\subsection{Match with Partial Overlap (\mpa)} \label{sec:mpa}
Exact Match might be too strict. We allow matches between error spans that overlap by at least $\tau$ characters by defining the set of one-to-one partial-overlap matchings:
\begin{equation*}
    \mathcal{M}_{\mpa}^{\tau} = \left\{ M \in \mathcal{M} \ \middle|\ 
    \forall (\hyp, s) \in M: | \hyp \cap s | \geq \tau
    \right\}
\end{equation*}
Then, given any $M \in \mathcal{M}_{\mpa}^{\tau}$, we define $P_{\mpa}$, $R_{\mpa}$, and $F_{\mpa}$ as in Equations~\ref{eq:em_precision}, \ref{eq:em_recall}, and \ref{eq:em_fscore}, and select $M^{\tau, *}_{\mpa}$ as the one that maximizes $F_{\mpa}$.

Returning to the example in Figure~\ref{fig:example1}, with $\tau=1$, then $|M_{\mpa}^{1, *}| = |\hyps| = |S| = 2$, as both hypothesis spans ``The quick'' and ``fox'' overlap with ground-truth spans by at least one character. As a consequence, $P_{\mpa} = R_{\mpa} = F_{\mpa} = 1$.

\subsection{Match with Partial Overlap and Partial Credit (\mpp)} \label{sec:mpp}
Both exact match and match with partial overlap might be too rigid: they count one true positive when two spans match, and nothing at all when spans go unmatched. Another option is to give partial credit when two spans overlap by a subset of characters. 

We define $\mathcal{M}_{\mpp} = \mathcal{M}_{\mpa}\big|_{\tau = 1}$ as the set of one-to-one partial-overlap matchings where $\tau=1$. Given any $M \in \mathcal{M}_{\mpp}$, one option is to define precision and recall as follows:\footnote{Throughout the paper, edge cases of precision and recall ($\hyps=\emptyset$ or $S=\emptyset$) are handled as in Equations~\ref{eq:em_precision} and \ref{eq:em_recall}.}
\begin{align}
\mathrm{P}_{\approx\textsc{w25}} &=
\frac{\sum_{(\hyp, s)\in M} | \hyp \cap s |}{\sum_{\hyp \in \hyps} | \hyp |} \\
\mathrm{R}_{\approx\textsc{w25}} &=
\frac{\sum_{(\hyp, s)\in M} | \hyp \cap s |}{\sum_{s \in S} |s|}
\end{align}
We dub this strategy $\approx$\textsc{w25} because this is very similar to the strategy adopted by \citet{lavie-etal-2025-findings} at the WMT25 Automated Translation and Evaluation shared task, with the main difference that they do not enforce a one-to-one matching $M$ between hypothesis and ground-truth spans.\footnote{We define formally the implementation used at WMT25 in Appendix~\ref{apx:wmt25_measure}.} However, we argue that this strategy is unsuitable for computing precision and recall in MT error detection because longer error spans weigh more than shorter ones. Indeed, precision and recall are computed from character counts, with error spans contributing to the final values in proportion to their length. However, the human evaluation protocols typically used as the ground-truth in MT evaluation are Multidimensional Quality Metrics (MQM, \citealp{mqm-framework, freitag-etal-2021-experts}) and Error Span Annotation (ESA, \citealp{kocmi-etal-2024-error}). 
In both annotation protocols, each error span identifies a single translation error and gets assigned a severity level that is irrespective of its length. Thus, we argue that a meta-evaluation strategy that uses MQM or ESA annotations as the ground truth should treat error spans of different length in the same way.

We enforce this constraint by calculating character-based precision and recall for each pair of matched spans in isolation, rather than for a full translation at once, and then averaging across error spans. Formally, given $M \in \mathcal{M}_{\mpp}$, we define precision and recall for each pair $(\hyp, s) \in M$:
\begin{equation*}
    P_{\textsc{mpp}}(\hyp, s) = \frac{\left| \hyp \cap s \right|}{\left| \hyp \right|} \quad 
    R_{\textsc{mpp}}(\hyp, s) = \frac{\left| \hyp \cap s \right|}{\left| s \right|}
\end{equation*} 
Then, sample-level precision and recall are: 
\begin{align}
    P_{\textsc{mpp}} &= \frac{1}{ |\hyps |}\sum_{(\hyp, s)\in M} P_{\textsc{mpp}}(\hyp, s) \\
    R_{\textsc{mpp}} &= \frac{1}{| S |}\sum_{(\hyp, s)\in M} R_{\textsc{mpp}}(\hyp, s)
\end{align}
By averaging across error spans, we ensure that each span has the same weight in span-level precision and recall, while still assigning partial credit based on character counts. Finally, $F_{\mpp}$ is computed as the harmonic mean of $P_{\mpp}$ and $R_{\mpp}$, and the optimal $M^*_{\mpp} \in  \mathcal{M}_{\mpp}$ is the one-to-one matching that maximizes $F_{\mpp}$.

\subsection{Averaging Results Across Data Samples}

Above, we illustrated several ways to compute precision, recall, and $\fscore$ for the evaluation of a single translation. Typically, however, we are interested in measuring performance across all samples in a given test set, which can be achieved either by macro-averaging or micro-averaging results. Formally, given a set of translations $D = \{ \bs{x}^1, \bs{x}^2, ..., \bs{x}^N \}$, we want to measure the precision, recall, and $\fscore$ of auto-evaluators on $D$.

\paragraph{Macro-averaging.}
This strategy involves averaging the values of precision, recall, and $\fscore$ across all $\bs{x} \in D$. 

\paragraph{Micro-averaging.} 
This strategy involves collecting statistics (e.g., true positives, false positives, and false negatives) across all samples, to later calculate precision, recall, and $\fscore$ over the entire test set once.\footnote{For \mpp, micro-averaging involves averaging span-level precision and recall across all spans in the test set, rather than accumulating true positives, false positives, and false negatives. This follows from precision and recall being already defined as averages in \mpp's sample-level version.} This procedure is equivalent to concatenating all translations in the test set and computing precision, recall, and $\fscore$ over the resulting sets of hypothesis and ground truth error spans. Accordingly, we construct $\bs{x} = \bs{x}^1 || \bs{x}^2 || ... || \bs{x}^N$ -- where we use $||$ for string concatenation -- and compute precision, recall, and $\fscore$ using the $\hyps^{\bs{x}}$ and $S^{\bs{x}}$, i.e., the sets of all hypothesis and ground-truth error spans in $\bs{x}$.

\section{Methodology}
Unlike MT, where the ground truth is typically a reference translation produced by professional human translators, or MT evaluation, where the ground truth is given by human annotations of translation quality, in MT meta-evaluation the ground-truth ranking of auto-evaluators is unknown. This makes it challenging to determine which meta-evaluation strategy is most appropriate, since we do not know which ranking of evaluators such a strategy should produce. Moreover, different variants of precision, recall, and $\fscore$ yield auto-evaluator rankings that differ drastically from one another (Table~\ref{tab:wmt25-mqm-ranks}).

Addressing a similar issue in score-level MT meta-evaluation, \citet{perrella-etal-2024-guardians} introduced the concept of sentinel auto-evaluators,\footnote{\citet{perrella-etal-2024-guardians} refer to them as sentinel metrics but we use ``auto-evaluator'' in place of ``metric''.} i.e., auto-evaluators that serve as probes to identify pitfalls in the meta-evaluation process. Sentinel auto-evaluators are designed such that assumptions can be made about where they should rank under a fair meta-evaluation, enabling the identification of meta-evaluation measures that rank them incorrectly. In this work, we design span-level sentinel auto-evaluators to test the robustness of span-level meta-evaluation measures.

\subsection{Imprecise Sentinel Auto-Evaluators}
We extend the error spans detected by our auto-evaluators to test the robustness of different implementations of precision, recall, and $\fscore$ to span length. Extended error spans are centered on the original error spans, but are less precise in pinpointing the exact error location.

\subsection{Low-recall Sentinel Auto-Evaluators}
MT error detection is sparse, with many samples featuring no ground-truth error spans. In MQM test sets, the proportion of translations annotated with zero errors ranges from $18.9\%$ to $77\%$, with the highest value observed in the WMT24 \langpair{en}{es} test set (full statistics are reported in Table~\ref{tab:data-statistics}). One consequence of this sparsity is that, under macro-averaging, many samples feature $R=1$ because recall's denominator is $0$ in Equation~\ref{eq:em_recall}. We hypothesize that this phenomenon may lead macro-averaging to favor auto-evaluators that inflate precision, predicting fewer errors to maximize the chance of getting $F=1$ on samples with no ground-truth errors.

To test this hypothesis, we create low-recall sentinel auto-evaluators by randomly deleting some of the errors detected by auto-evaluators. Varying the probability with which we delete each error, we measure the robustness of different meta-metrics to task sparsity.

\begin{figure*}[ht]
    \centering

    \includegraphics[width=\linewidth]{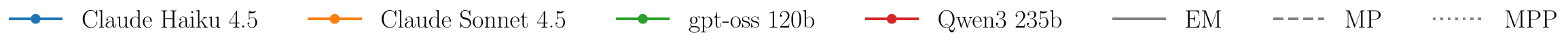}

    \medskip
    
    % First subfigure
    \centering
    \includegraphics[width=0.49\linewidth]{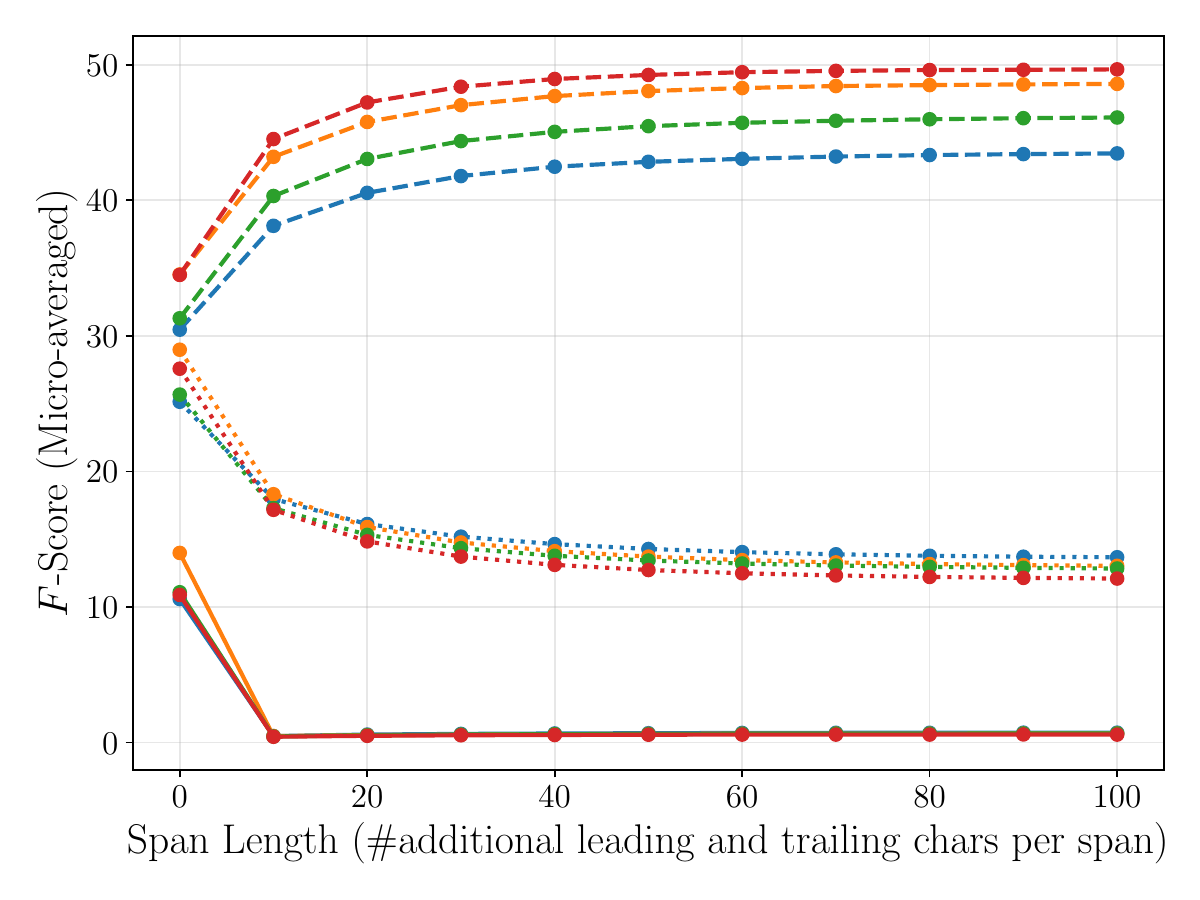}
    \hfill
    \includegraphics[width=0.49\linewidth]{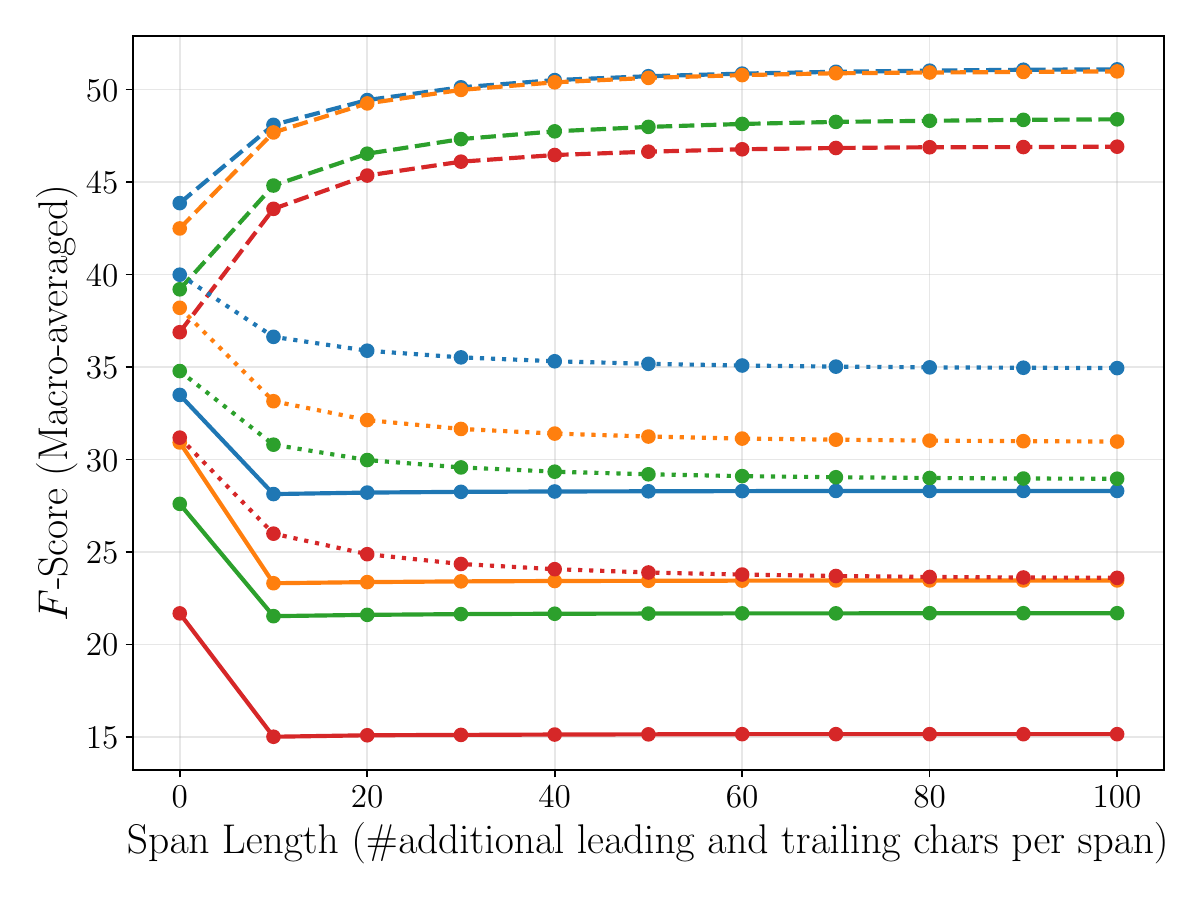}
    \caption{Performance of auto-evaluators and imprecise sentinels at varying span length, with micro (left) and macro (right) averaging, under all meta-metrics defined in Section~\ref{sec:span-level-meta-evaluation}. $x=0$ shows the performance of the base auto-evaluators before their spans were extended.}
    \label{fig:imprecise-results}
\end{figure*}

\section{Experimental Setup}
We demonstrate that some meta-evaluation strategies are unsuitable for MT error detection by measuring the performance of sentinel auto-evaluators alongside normal auto-evaluators.

\paragraph{Data} \label{sec:data}
We conduct our experiments on the concatenation of MQM test sets released from 2022 to 2024 \cite{freitag-etal-2022-results, freitag-etal-2023-results, riley-etal-2024-finding, freitag-etal-2024-llms}. Table~\ref{tab:data-statistics} in Appendix~\ref{apx:data-and-models} reports statistics about the test sets and language directions employed in this work.

\paragraph{Models}
Our auto-evaluators consist of four Large Language Models (LLMs) prompted for error detection: \sonnet, \haiku, \qwen, and \gptoss. This selection of models includes closed-source and open-weights models of varying sizes and families.\footnote{Implementation details are reported in Appendix~\ref{apx:implementation_details}.} 

\section{Results}
We measure the performance of our imprecise and low-recall sentinels alongside normal auto-evaluators under the implementations of precision, recall, and $\fscore$ illustrated in Section~\ref{sec:span-level-meta-evaluation}, i.e., \exm, \mpa, and \mpp, with micro- and macro-averaging. \mpa uses $\tau=1$, matching spans that overlap by $1$ or more characters. We chose the minimum value for $\tau$ for two main reasons: (1) it does not affect the generalization of our results while representing the most extreme case, which facilitates visualization of results, and (2) several previous works have used variants of \mpa with $\tau=1$ (Section~\ref{sec:previous-work}).

\begin{figure*}[ht]
    \centering

    \includegraphics[width=\linewidth]{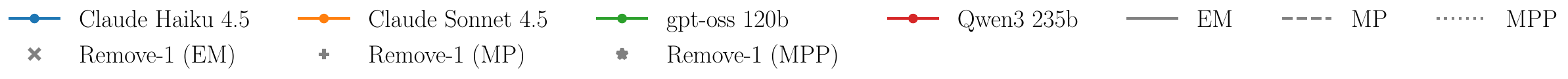}

    \medskip
    
    % First subfigure
    \centering
    \includegraphics[width=0.49\linewidth]{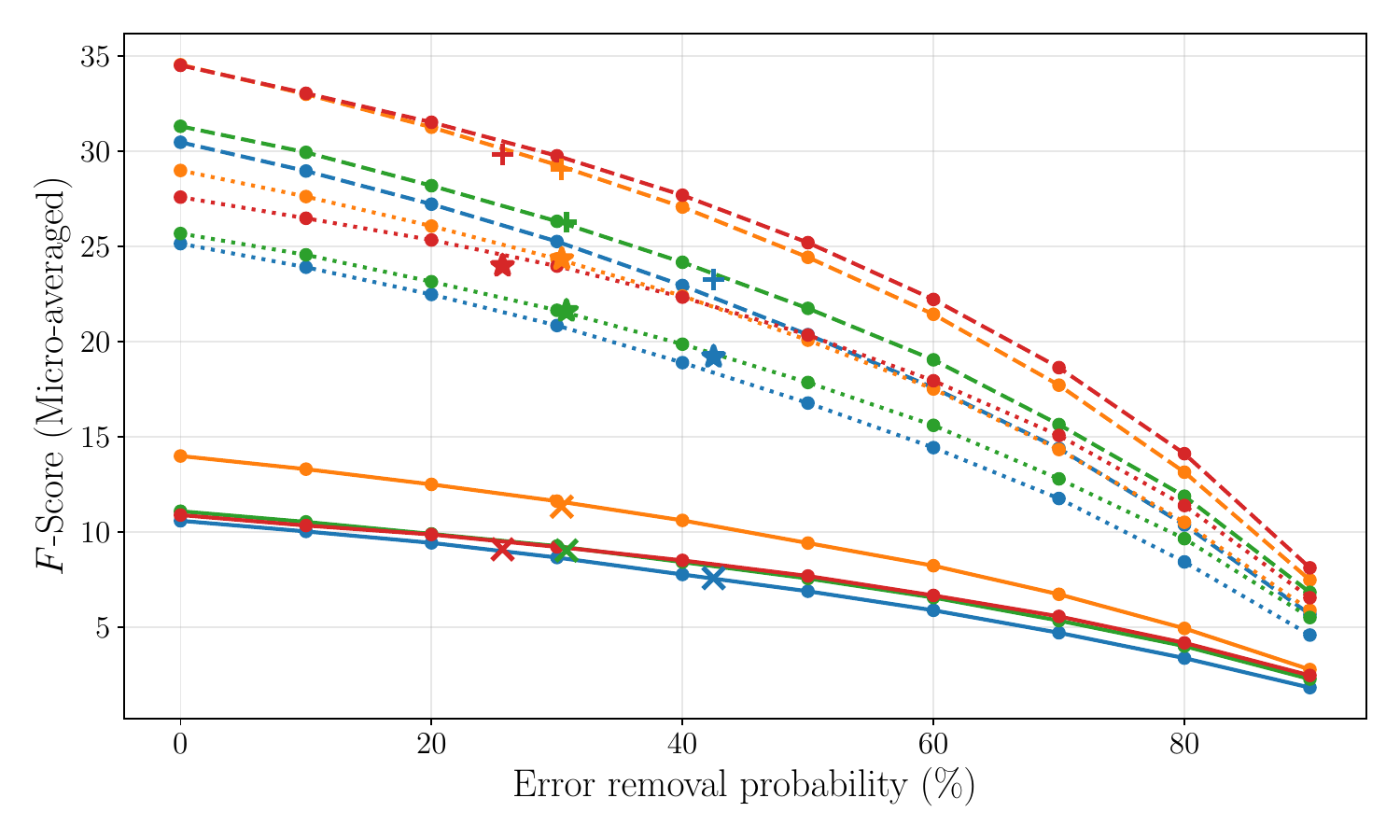}
    \hfill
    \includegraphics[width=0.49\linewidth]{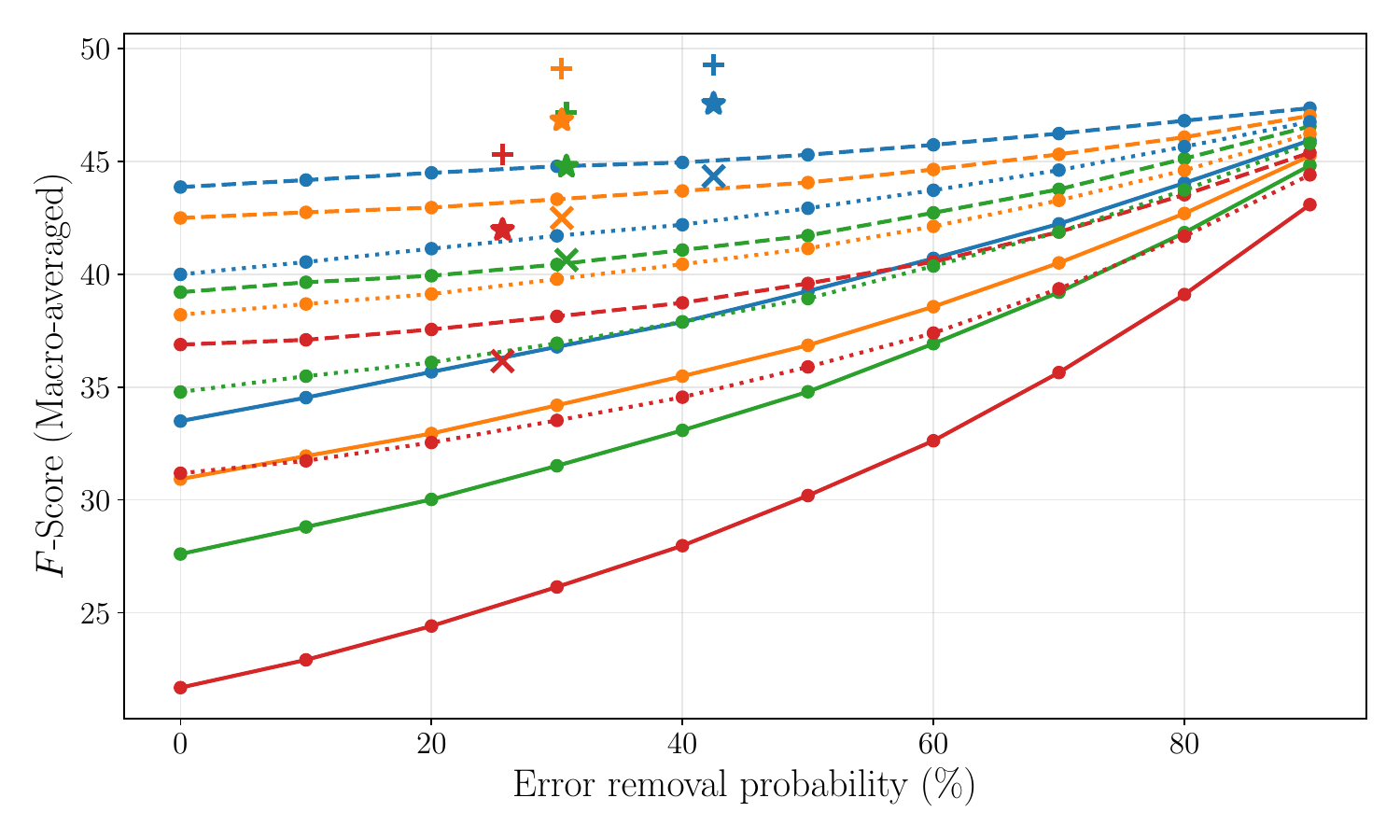}
    \caption{Performance of auto-evaluators and low-recall sentinels at varying error removal probabilities, with micro (left) and macro (right) averaging, under all meta-metrics defined in Section~\ref{sec:span-level-meta-evaluation}. $x=0$ shows the performance of the base auto-evaluators before any span was removed from their predictions.}
    \label{fig:low-recall-results}
\end{figure*}

\subsection{Performance of Imprecise Sentinels} \label{sec:imprecise-results}
We extend the error spans produced by our auto-evaluators by progressively adding leading and trailing characters, and report results under \exm, \mpa, and \mpp, using micro- (Figure~\ref{fig:imprecise-results}, left) and macro-averaging (Figure~\ref{fig:imprecise-results}, right).

Performance under \exm and \mpp appropriately decreases as span length increases, whereas performance under \mpa improves. By assigning full credit to spans whose overlap exceeds $\tau$ characters, \mpa inflates the precision, recall, and $\fscore$ of auto-evaluators that produce longer spans, as longer spans are more likely to yield a match. As a result, \mpa is unsuitable for span-level MT meta-evaluation because it can be easily gamed by returning artificially long error spans. Even in the absence of deliberate metric manipulation, our results show that \mpa is biased toward auto-evaluators that produce longer spans, thereby confounding evaluations.

Additionally, we observe that $\fscore_{\exm}$ drops abruptly to near zero for all auto-evaluators when span length is increased by even a small number of characters. This behavior is expected, as \exm requires exact span overlap. Nonetheless, it further highlights that \exm may be too strict for error detection evaluation, an inherently noisy task in which human annotators often disagree on the precise locations and boundaries of translation errors.\footnote{We refer the reader to Section~\ref{sec:human-parity}, where we measure agreement between human evaluators.}

\subsection{Performance of Low-Recall Sentinels} \label{sec:low-recall-results}
We delete errors detected by our auto-evaluators uniformly at random with progressively higher probability and report results under \exm, \mpa, and \mpp, with micro- (Figure~\ref{fig:low-recall-results}, left) and macro-averaging (Figure~\ref{fig:low-recall-results}, right).

Despite differing solely in how results are averaged, micro- and macro-averaging produce drastically different outcomes. Specifically, under micro-averaging, performance decreases as the removal probability increases, whereas under macro-averaging it increases, highlighting its lack of robustness to task sparsity. Indeed, because many samples contain no ground-truth errors (Table~\ref{tab:data-statistics}), low-recall sentinels yield higher $\fscore$ under macro-averaging by maximizing the number of samples assigned with zero errors.

We further exploit this limitation of macro-averaging by deleting all errors detected by an auto-evaluator in samples where it detected $\leq 1$ errors, and denote the $\fscore$ of these modified sentinel auto-evaluators with ``Remove-1'' in Figure~\ref{fig:low-recall-results}. For each sample, these sentinels return either $0$ or $\geq 2$ error spans. As shown in the figure, Remove-1 sentinels further increase macro-averaged $\fscore$. In contrast, micro-averaging is robust to task sparsity: $\fscore$ progressively decreases as the error removal probability increases, and ``Remove-1'' sentinels do not achieve inflated scores, as their $\fscore$ lies close to the corresponding auto-evaluator's curve.

\subsection{Impact on Auto-Evaluator Ranking}
Previous results establish that \mpa and macro-averaging are biased to prefer auto-evaluators that return longer error spans and under-detect translation errors, respectively. Additionally, in previous sections we also argued that \exm might be too strict for span-level MT meta-evaluation, and that the approach adopted in the WMT 2025 Automated Translation Evaluation Systems shared task -- which we dub \textsc{w25} -- is misaligned with the MQM and ESA annotations that serve as ground truth in MT evaluation (Section~\ref{sec:mpp}). In this section, we ask whether these factors also translate into differences in auto-evaluator rankings in realistic evaluation scenarios and, if so, to what extent. To investigate this question, we compute the ranking of auto-evaluators on the MQM split of the WMT 2025 shared task using all the metrics discussed above, including \textsc{w25}, which we discuss in more detail and formally define in Appendix~\ref{apx:wmt25_measure}. We present the resulting auto-evaluator rankings in Table~\ref{tab:wmt25-mqm-ranks}, whereas Table~\ref{tab:wmt25-mqm-scores} in Appendix~\ref{apx:additional-results} reports the corresponding $F$-scores.

The most noticeable difference is between the micro- and macro-averaged results: the ranking is almost entirely reversed, with auto-evaluators moving from the top to the bottom of the ranking, and vice versa, when switching from micro- to macro-averaging. Notably, AutoLQA41.sec moves from last overall to first overall, while \qwen moves from first overall to near the bottom of the ranking. As shown in Table~\ref{tab:wmt25-stats-mqm}, AutoLQA41.sec is the model that identifies the fewest translation errors, while \qwen is among those that identify the most. This further supports our finding that macro-averaging is biased toward low-recall auto-evaluators. Moreover, these results provide an indication of how strongly this bias can affect the final auto-evaluator rankings.

Substantial ranking differences also emerge when comparing measures that use the same averaging strategy. Focusing on micro-averaging, the positions of individual auto-evaluators vary considerably across measures in several cases. For example, \xcometxl ranks 12th, 6th, 8th, and 3rd under \exm, \mpa, \mpp, and \textsc{w25}, respectively. Similarly, AIP.sec ranks 1st, 7th, 6th, and 12th. These large differences in auto-evaluator rankings underscore the importance of establishing a robust approach to span-level meta-evaluation: seemingly small methodological choices can not only introduce evaluation biases, but these biases can also substantially affect the resulting auto-evaluator rankings when the measures are applied in realistic evaluation settings.

\section{Approaches Used in Previous Work} \label{sec:previous-work}
Previous work has adopted disparate approaches. In the 2019 and 2020 editions of the WMT Quality Estimation shared tasks \cite{fonseca-etal-2019-findings, specia-etal-2020-findings-wmt}, the organizers used versions of span-level precision, recall, and $\fscore$ similar to \mpp (we refer to this measure as \textsc{w}$19$ and formally define it in Appendix~\ref{apx:wmt19_and_20}). However, they compute the final $\fscore$ via macro-averaging, which favors low-recall auto-evaluators, as shown in Section~\ref{sec:low-recall-results}. In the WMT Quality Estimation shared tasks in 2023 and 2024 \cite{blain-etal-2023-findings, zerva-etal-2024-findings}, as well as in the WMT shared task on Automated Translation Quality Evaluation Systems in 2025 \cite{lavie-etal-2025-findings}, the organizers adopted different evaluation strategies, which we denote by \textsc{w23} and \textsc{w25}, and define formally in Appendix~\ref{apx:wmt25_measure}. However, both strategies assign greater weight to longer errors than shorter ones, which is not aligned with how severity is defined in the ESA and MQM protocols. Furthermore, \citet{lavie-etal-2025-findings} report findings based on both micro- and macro-averaged results.

Beyond the WMT shared tasks, other previous work has adopted one or the other approach. \citet{perrella-etal-2022-matese} evaluate their \matese metrics using the Span Hit Metrics -- roughly equivalent to \mpa with $\tau=1$. Similarly, \citet{kocmi-etal-2024-error} measure inter-annotator agreement between different human evaluation protocols in terms of how much protocol A covers protocol B, a measure similar to $R_{\mpa}$ with $\tau=1$. However, as shown in Section~\ref{sec:imprecise-results}, \mpa is biased toward auto-evaluators (or human annotators) that produce longer spans. \citet{guerreiro-etal-2024-xcomet} evaluate XCOMET using $\textsc{w}23$. \citet{fernandes-etal-2023-devil} evaluate AutoMQM by defining custom meta-metrics similar to $\textsc{w}23$, where span overlap is computed based on word counts rather than character counts. Finally, \citet{kasner2026llmsspanannotatorscomparative} use the measures defined by \citet{da-san-martino-etal-2019-fine} for propaganda detection, which closely resemble \textsc{w}$19$, but they do not clarify whether micro- or macro-averaging is used to derive corpus-level statistics.

\subsection{Summary of Findings}

\begin{itemize}
    \item ``Match with partial overlap'' (\mpa) favors auto-evaluators that produce longer error spans.

    \item Macro-averaging across data samples favors auto-evaluators that under-detect translation errors. 

    \item ``Exact match'' (\exm) may be too strict for span-level MT meta-evaluation.

    \item Auto-evaluator rankings change drastically when switching from micro- to macro-averaging, as well as across different implementations of span-level precision, recall, and $\fscore$.

    \item \textbf{We recommend ``match with partial overlap and partial credit'' (\mpp) combined with micro-averaging}. Among the measures studied, this approach is the most robust to variations in span length and task sparsity, while remaining flexible enough to account for partially detected errors.
\end{itemize}

\section{Assessing the State of the Art}
Having established \mpp with micro averaging as the recommended span-level MT meta-evaluation strategy, we use it for two research purposes: In Section~\ref{sec:wmt25-ranking}, we evaluate auto-evaluators and rank them according to their capabilities; In Section~\ref{sec:human-parity}, we assess the state of progress in MT error detection by measuring the gap between auto-evaluators and human annotators. 

\begin{table}[t]
\centering
\small 

\begin{tabular}{lrrr}
\toprule
\textbf{Metric} & \multicolumn{1}{c}{$P$} & \multicolumn{1}{c}{$R$} & \multicolumn{1}{c}{$F$1} \\
\midrule
\cellcolor{highlight}Qwen3 235b & \cellcolor{highlight} $17.36$ & \cellcolor{highlight} $31.45$ & \cellcolor{highlight} $22.37$ \\
GemSpanEval-QE.sec & $16.33$ & $28.61$ & $20.60$ \\
\cellcolor{highlight}Claude Sonnet-4.5 & \cellcolor{highlight} $21.47$ & \cellcolor{highlight} $18.58$ & \cellcolor{highlight} $19.92$ \\
GemSpanEval.pri & $16.60$ & $25.34$ & $19.84$ \\
XCOMET-XXL.bas & $16.38$ & $21.39$ & $18.33$ \\
AIP.sec & $37.34$ & $11.99$ & $18.16$ \\
\cellcolor{highlight}Claude Haiku 4.5 & \cellcolor{highlight} $21.95$ & \cellcolor{highlight} $14.82$ & \cellcolor{highlight} $17.69$ \\
XCOMET-XL.bas & $15.46$ & $21.10$ & $17.47$ \\
\cellcolor{highlight}gpt-oss 120b & \cellcolor{highlight} $19.57$ & \cellcolor{highlight} $15.78$ & \cellcolor{highlight} $17.45$ \\
AIP.pri & $35.18$ & $10.30$ & $15.93$ \\
AutoLQA.pri & $16.00$ & $14.99$ & $14.95$ \\
AutoLQAESA.sec & $16.46$ & $13.00$ & $14.01$ \\
AutoLQA41.sec & $24.42$ & $5.79$ & $9.30$ \\
\bottomrule
\end{tabular}

\caption{\mpp with micro averaging on the MQM split of WMT25. We run the auto-evaluators highlighted in grey, the others are submissions to WMT25 \cite{lavie-etal-2025-findings}.}
\label{tab:wmt25-full-results-mqm}
\end{table}

\subsection{WMT25 Ranking} \label{sec:wmt25-ranking}
Table~\ref{tab:wmt25-full-results-mqm} presents the auto-evaluator ranking on the MQM split of WMT25 -- including the English-to-Korean and Japanese-to-Chinese translation directions -- using \mpp with micro-averaging. We also compute the auto-evaluator ranking on the ESA split of WMT25 and report results in Table~\ref{tab:wmt25-full-results-esa} in Appendix~\ref{apx:additional-results}. Results are aggregated across translation directions by averaging language pair-specific precision, recall, and $\fscore$.\footnote{Unlike WMT25, we do not apply any severity penalty, meaning that matches between errors of any severity are treated equally. Our work focuses on evaluating error detection capabilities, and our meta-evaluation strategies are orthogonal to the application of severity penalties.}

The first and third positions are obtained by \qwen and \sonnet, respectively, which are based on our reference-less, zero-shot, LLM-as-a-Judge approach illustrated in Appendix~\ref{apx:implementation_details}. The second and fourth positions are achieved by GemSpanEval models, which are based on the Gemma-3 27b model \cite{gemmateam2025gemma3technicalreport} fine-tuned for error detection \cite{juraska-etal-2025-metricx}. Interestingly, GemSpanEval-QE, which operates without relying on reference translations, slightly outperforms GemSpanEval.

We also note that results differ substantially when ESA is used as the ground-truth, both in terms of ranking and absolute scores (Table~\ref{tab:wmt25-full-results-esa}). In particular, we observe that average $F$-scores are substantially lower with ESA. While this might be attributed to the different translation directions covered by the two test sets, it might also stem from the fact that ESA annotators have been shown to detect fewer errors compared to the MQM annotations conducted at WMT \cite{kocmi-etal-2024-error}. Indeed, the average precision of auto-evaluators is $21.57$ for MQM error detection, whereas it is only $10.19$ with ESA; conversely, average recall is higher with ESA than with MQM ($20.55$ vs $16.95$).\footnote{We compute averages excluding the auto-evaluators run by us because they were not evaluated on the ESA test set.}

% 6-model gradient: rank 1 (best) -> rank 6 (worst)
\newcommand{\rankcolorSix}[1]{%
  \ifcase#1\or
    SeaGreen3!100!Firebrick3!50\or % 1
    SeaGreen3!90!Firebrick3!50\or  % 2
    SeaGreen3!80!Firebrick3!50\or  % 3
    SeaGreen3!70!Firebrick3!50\or  % 4
    SeaGreen3!60!Firebrick3!50\or  % 5
    SeaGreen3!50!Firebrick3!50     % 6
  \fi
}
\newcommand{\fcell}[2]{\cellcolor{\rankcolorSix{#1}}$#2$} % #1=rank, #2=value

\begin{table*}[t]
\centering
\small

\begin{tabular}{lrrrrrrrrrrrr}
\toprule
\textbf{Metric} & \multicolumn{3}{c}{\langpair{en}{de} (2022)} & \multicolumn{3}{c}{\langpair{en}{zh} (2022)} & \multicolumn{3}{c}{\langpair{en}{de} (2023)} & \multicolumn{3}{c}{\langpair{zh}{en} (2023)} \\
 & \multicolumn{1}{c}{P} & \multicolumn{1}{c}{R} & \multicolumn{1}{c}{$F$1} & \multicolumn{1}{c}{P} & \multicolumn{1}{c}{R} & \multicolumn{1}{c}{$F$1} & \multicolumn{1}{c}{P} & \multicolumn{1}{c}{R} & \multicolumn{1}{c}{$F$1} & \multicolumn{1}{c}{P} & \multicolumn{1}{c}{R} & \multicolumn{1}{c}{$F$1} \\
\midrule
MQM \#1      & $42.66$ & $44.62$ & \fcell{1}{43.62} & $29.88$ & $25.49$ & \fcell{3}{27.51} & $38.04$ & $35.51$ & \fcell{2}{36.73} & $40.17$ & $39.48$ & \fcell{3}{39.82} \\
MQM \#2      & $39.22$ & $43.56$ & \fcell{2}{41.28} & $35.95$ & $25.68$ & \fcell{1}{29.96} & $39.02$ & $37.47$ & \fcell{1}{38.23} & $44.57$ & $39.82$ & \fcell{1}{42.06} \\
Sonnet 4.5   & $32.61$ & $29.75$ & \fcell{3}{31.12} & $34.31$ & $21.57$ & \fcell{4}{26.49} & $38.92$ & $27.80$ & \fcell{3}{32.44} & $48.82$ & $33.94$ & \fcell{2}{40.04} \\
Qwen3 235b   & $23.62$ & $38.87$ & \fcell{4}{29.39} & $28.88$ & $30.92$ & \fcell{2}{29.87} & $31.18$ & $30.41$ & \fcell{4}{30.79} & $40.12$ & $39.50$ & \fcell{4}{39.81} \\
gpt-oss 120b & $24.08$ & $31.04$ & \fcell{5}{27.12} & $28.89$ & $21.42$ & \fcell{5}{24.60} & $32.33$ & $24.60$ & \fcell{5}{27.94} & $44.55$ & $29.10$ & \fcell{5}{35.20} \\
Haiku 4.5    & $30.07$ & $19.89$ & \fcell{6}{23.94} & $29.88$ & $15.15$ & \fcell{6}{20.11} & $37.79$ & $18.54$ & \fcell{6}{24.87} & $50.25$ & $25.66$ & \fcell{6}{33.97} \\
\bottomrule
\end{tabular}

\caption{Results on the \langpair{en}{de} MQM test set from WMT 2022 \cite{kocmi-etal-2022-findings, freitag-etal-2022-results}, the \langpair{en}{zh} MQM annotations released by \citet{riley-etal-2024-finding}, and the \langpair{en}{de} and \langpair{zh}{en} MQM test sets from WMT 2023 \cite{kocmi-etal-2023-findings, freitag-etal-2023-results}. ``MQM \#1'' and ``MQM \#2'' are human evaluators.}

\label{tab:human-parity-wmt22-wmt23}
\end{table*}

\subsection{State of Progress} \label{sec:human-parity}
To get a sense of the state of progress in MT error detection, we measure the performance gap between human annotators and auto-evaluators, using inter-annotator agreement as a reference for human performance. We adopt the same approach as \citet{proietti-etal-2025-machine}, who assess whether human parity has been achieved in score-level MT evaluation. Specifically, we evaluate humans and auto-evaluators jointly under the same evaluation measure (\mpp with micro-averaging). 

We use the MQM test sets released at the WMT Metrics Shared Task in 2022 and 2023 \cite{freitag-etal-2022-results, freitag-etal-2023-results}, alongside the MQM test set released by \citet{riley-etal-2024-finding}. Translations in these test sets were each annotated three times by distinct human evaluators. This redundancy allows us to select one set of annotations as the ground-truth and use the others to estimate human performance.\footnote{We note that each human evaluator does not correspond to a single annotator, but rather to the combination of multiple annotators, because WMT evaluation campaigns typically collect annotations from a pool of annotators. Nonetheless, each sample is annotated by three \textit{distinct} annotators.} For this experiment, we cannot use the WMT24 \cite{kocmi-etal-2024-findings, freitag-etal-2024-llms} and WMT25 test sets \cite{kocmi-etal-2025-findings, lavie-etal-2025-findings}. In the 2024 ESA and MQM test sets, as well as in the WMT25 MQM test set, each translation was annotated only once. Regarding the ESA split of WMT25, although many translations were annotated multiple times, there is no guarantee that the annotations were produced by distinct annotators, which would inflate our estimate of human reference performance.

We report results in Table~\ref{tab:human-parity-wmt22-wmt23}. On average, MQM human evaluators have higher performance than auto-evaluators, especially for \langpair{en}{de}, where the $F$-scores achieved by humans are as much as $10$ points higher than those of auto-evaluators. Only twice is a human evaluator surpassed by automatic systems: MQM \#1 ranks below \qwen in \langpair{en}{zh}, and below \sonnet in \langpair{zh}{en}. In contrast, MQM \#2 consistently ranks above auto-evaluators; only in \langpair{en}{zh} \qwen achieves an $\fscore$ that matches its performance. 

These results paint a different picture from the findings of \citet{proietti-etal-2025-machine}. They measure human and auto-evaluator performance at the score level, using the same test sets and human annotations used in this work, and find that human evaluators are consistently matched or surpassed by automatic systems, raising concerns about our ability to measure progress in score-level MT evaluation. In their discussion, they attribute these findings to limitations in meta-evaluation measures, annotation quality, and benchmark difficulty, urging the research community to address these issues to ensure progress in MT evaluation remains measurable. While some of these issues also extend to span-level meta-evaluation, we hypothesize that, by factoring in error location, the span-level setting may enable more precise estimates of auto-evaluator capabilities, thereby increasing meta-evaluation resolution. For example, consider an auto-evaluator that incorrectly detects an error at one location while simultaneously missing a true error at another location. The MQM-based score assigned to the translation would remain unchanged compared to the case where the correct error had been detected instead. This additional noise in score-level meta-evaluation may partly explain why human evaluators are consistently matched by auto-evaluators, whereas they still rank higher at the span-level.

\section{Conclusions}
This work investigates strategies for reliably assessing the performance of machine translation error detection. We evaluate the robustness of three versions of span-level precision, recall, and $\fscore$, both micro- and macro-averaging results across data samples. Our results show that different measures yield substantially different rankings of auto-evaluators and that several widely used approaches introduce systematic unfairness into the meta-evaluation, favoring auto-evaluators that produce longer error spans or that under-detect translation errors. Based on these findings, we recommend ``match with partial overlap and partial credit'' (\mpp) combined with micro-averaging as a robust strategy for span-level machine translation meta-evaluation. Finally, using \mpp, we establish the state of the art in MT error detection and measure the gap between state-of-the-art auto-evaluators and human performance.

\section*{Limitations}

\paragraph{Characters vs words vs tokens}
Both ``match with partial overlap'' (\mpa) and ``match with partial overlap and partial credit'' (\mpp) rely on character-based overlap to derive a one-to-one matching. Specifically, two spans can match only if they overlap by at least one character (or $\tau$ characters, for \mpa). Similarly, the partial credit assigned by \mpp to valid matches corresponds to the proportion of overlapping characters. We selected characters as the unit of overlap because this makes our measures tokenization-agnostic while still working well with scripts that do not rely on spaces to separate words, such as Chinese or Japanese. A consequence of this choice is that long words contribute more to precision and recall than short ones. While we believe that the impact of this phenomenon is negligible, our work does not investigate whether alternative choices would significantly change results. 

\paragraph{Severity weighting}
This work does not investigate weighting precision and recall according to error severity, which is arguably orthogonal to our discussion of evaluation measures for error detection. Recent editions of the WMT Quality Estimation shared task \cite{blain-etal-2023-findings, zerva-etal-2024-findings} and the Automated Translation Evaluation Systems shared task \cite{lavie-etal-2025-findings} apply a $0.5$ penalty to error spans with mismatched severity. However, as also noted by \citet{lyu2025minimumbayesriskdecoding}, this strategy penalizes precision and recall when predicted and ground-truth severities differ, but does not account for the fact that predicting a \minor error where there is no ground-truth error is less severe than predicting a \major error. \citet{aur-etal-2026-exploring} adapt \mpp to incorporate a severity-weighting scheme that addresses this issue: severity mismatches remain penalized, while errors of different severities contribute to the final \fscore in proportion to their respective weights. Our implementation at \url{https://github.com/sted19/span-mt-metrics-eval} includes options for using \mpp with this severity weighting.

\section*{Acknowledgments}

The authors thank Pablo Castells, Ronald Denaux, Michael Denkowski, Javier Ferrando, and Esther López Ramos for their valuable feedback.

\bibliography{custom,anthology-1,anthology-2}

\clearpage
\appendix

\section{Data and Models} 
\label{apx:data-and-models}
Table~\ref{tab:data-statistics} reports statistics for the MQM and ESA test sets employed in this work. Table~\ref{tab:citations} lists the WMT25 submissions used in this work and attributes them to their corresponding research papers.

\begin{table}[t]
    \small
    \centering
    \resizebox{\columnwidth}{!}{
        \begin{tabular}{llrrr}
             \toprule
             \multicolumn{1}{l}{\small Year} & \multicolumn{1}{l}{\small \makecell{Language\\Pair}} & \multicolumn{1}{c}{\small \#Samples} & \multicolumn{1}{c}{\small \#Errors} & \multicolumn{1}{c}{\small \makecell{\%Samples\\w/o Errors}} \\
             \midrule
             
             \multirow{2}{*}{2022} & \langpair{en}{de} & $21,040$ & $16,074$ & $53.5\%$ \\
             & \langpair{en}{zh} & $49,558$ & $29,558$ & $48.6\%$ \\
             \cmidrule(lr){1-5}
             
             \multirow{2}{*}{2023} & \langpair{en}{de} & $5,980$ & $18,698$ & $26.9\%$ \\
             & \langpair{zh}{en} & $18,829$ & $43,767$ & $18.9\%$ \\
             \cmidrule(lr){1-5}
             
             \multirow{3}{*}{2024} & \langpair{en}{de} & $9,253$ & $11,650$ & $48.3\%$ \\
             & \langpair{en}{es} & $9,345$ & $4,077$ & $77.4\%$  \\
             & \langpair{ja}{zh} & $8,400$ & $8,378$ & $51.0\%$ \\
             \cmidrule(lr){1-5}
             
             \multirow{2}{*}{\makecell[l]{2025\\MQM}} & \langpair{en}{ko} & $3,474$ & $7,706$ & $38.6\%$ \\
             & \langpair{ja}{zh} & $3,906$ & $8,591$ & $27.0\%$ \\
             \cmidrule(lr){1-5}
             
             \multirow{14}{*}{\makecell[l]{2025\\ESA}} & \langpair{cs}{de} & $9,140$ & $17,238$ & $59.8\%$ \\
             & \langpair{cs}{uk} & $7,290$ & $6,561$ & $67.0\%$ \\
             & \langpair{en}{ar} & $6,640$ & $6,069$ & $53.0\%$ \\
             & \langpair{en}{bho} & $11,620$ & $12,681$ & $74.9\%$ \\
             & \langpair{en}{cs} & $7,296$ & $12,532$ & $64.1\%$ \\
             & \langpair{en}{et} & $6,308$ & $12,399$ & $57.4\%$ \\
             & \langpair{en}{is} & $5,976$ & $17,301$ & $45.4\%$ \\
             & \langpair{en}{it} & $5,976$ & $7,620$ & $55.3\%$ \\
             & \langpair{en}{ja} & $5,976$ & $3,968$ & $73.3\%$ \\
             & \langpair{en}{mas} & $8,964$ & $4,106$ & $63.2\%$ \\
             & \langpair{en}{ru} & $5,976$ & $9,188$ & $55.7\%$ \\
             & \langpair{en}{sr} & $5,976$ & $12,048$ & $52.1\%$ \\
             & \langpair{en}{uk} & $5,976$ & $3,363$ & $72.6\%$ \\
             & \langpair{en}{zh} & $5,976$ & $6,695$ & $68.6\%$ \\
            \bottomrule
        \end{tabular}
    }
    \caption{Statistics of the test sets used in this work. These test sets have been sourced from several previous works: the WMT Metrics Shared Task editions from 2022 to 2024 \cite{freitag-etal-2022-results, freitag-etal-2023-results, freitag-etal-2024-llms}, the study from \citet{riley-etal-2024-finding} regarding finding replicable human evaluation techniques, the WMT25 General Machine Translation shared task \cite{kocmi-etal-2025-findings}, and the WMT25 shared task on Automated Translation Evaluation Systems \cite{lavie-etal-2025-findings}.}
    \label{tab:data-statistics}
\end{table}
\begin{table}[ht]
\centering
\small
\begin{tabular}{ll}
\toprule
Auto-evaluator & Research Paper \\
\midrule
GemSpanEval & \cite{juraska-etal-2025-metricx} \\
GemSpanEval-QE & \cite{juraska-etal-2025-metricx} \\
XCOMET-XL & \cite{guerreiro-etal-2024-xcomet} \\
XCOMET-XXL & \cite{guerreiro-etal-2024-xcomet} \\
AIP (pri) & \cite{yeom-etal-2025-tagged} \\
AIP (sec) & \cite{yeom-etal-2025-tagged} \\
AutoLQA & \cite{hrabal-etal-2025-cuni} \\
AutoLQAESA & \cite{hrabal-etal-2025-cuni} \\
AutoLQA41 & \cite{hrabal-etal-2025-cuni} \\
\bottomrule
\end{tabular}
\caption{Research papers associated with the WMT25 submissions and baselines used in this work.}
\label{tab:citations}
\end{table}

\section{LLM-as-a-Judge Implementation Details} \label{apx:implementation_details}

Our auto-evaluators consist of four Large Language Models (LLMs) prompted for error detection:
\begin{itemize}
    \item \textbf{\sonnet} is a closed-source thinking LLM developed by Anthropic, optimized for agents, coding, and computer use.\footnote{\url{https://www.anthropic.com/claude/sonnet}.}
    \item \textbf{\haiku} is a cost-efficient thinking model in the Claude family.\footnote{\url{https://www.anthropic.com/claude/haiku}.}
    \item \textbf{\qwen} is an open-weight Mixture-of-Experts model from the Qwen3 model family \cite{yang2025qwen3technicalreport}. It features 235b parameters, with only 22b parameters activated per token during a forward pass. This model incorporates both thinking and non-thinking modes.
    \item \textbf{\gptoss} is an open-weight, Mixture-of-Experts, thinking model from the GPT model family \cite{openai2025gptoss120bgptoss20bmodel}. It features 116.8b parameters, with only 5.13b parameters activated per token during a forward pass. 
\end{itemize}

We instruct these LLMs using the prompt reported in Table~\ref{tab:error-detection-prompt}. Such a prompt illustrates precisely the error detection task, providing to the model:
\begin{itemize}
    \item \textbf{Task Guidelines}: high-level instructions on how translation errors should be identified;
    \item \textbf{Error Typology}: the MQM error typology used by \citet{freitag-etal-2021-experts};
    \item \textbf{Severity Levels}: the allowed severity levels;
    \item \textbf{Output Annotation Format}: the JSON output annotation format expected;
    \item \textbf{Task Instructions}: step-by-step instructions guiding the models to conduct the error detection task.
\end{itemize}
\texttt{\{final\_instruction\}} is assigned one out of two different values depending on whether the underlying LLM uses reasoning or not:
\begin{enumerate}
    \item \textbf{Reasoning:} \texttt{Execute these steps sequentially. Ensure you complete all steps: error identification, individual error analysis (including reasoning, categorization, and verification), and final annotation generation. Your output outside thinking tags must contain only the final json annotation.}

    \item \textbf{Non-reasoning:} \texttt{Execute these steps sequentially. Ensure your output shows your reasoning for all steps: error identification, individual error analysis (including reasoning, categorization, and verification), and final annotation generation.}
\end{enumerate}
Asking non-reasoning models to show their reasoning in the output, we elicit some reasoning behavior, preventing them from outputting the final JSON directly. Reasoning models are \gptoss, \sonnet, and \haiku, while \qwen is used without reasoning.

We collect our evaluations using the AWS Bedrock service.\footnote{\url{https://aws.amazon.com/bedrock/}.} We sample from LLMs using the following generation parameters, leaving the rest as per the Bedrock default:

\begin{center}
\begin{tabular}{l|r}
    \toprule
     max\_new\_tokens & $32768$ \\
     reasoning\_effort & \texttt{medium} \\
     reasoning\_budget & $4096$ \\
     \bottomrule
\end{tabular}
\end{center}

\section{Measures Used at WMT} \label{apx:wmt-measures}
WMT editions have employed different variants of span-level precision, recall, and \fscore to measure error detection performance. In this section, we formally define these measures and highlight their differences from those introduced in Section~\ref{sec:span-level-meta-evaluation}. We use the same notation as in Section~\ref{sec:span-level-meta-evaluation}. Importantly, as in the rest of the paper, we ignore error severity.

\subsection{WMT 2019 and 2020} \label{apx:wmt19_and_20}
At the 2019 and 2020 editions of the WMT Quality Estimation shared tasks \cite{fonseca-etal-2019-findings, specia-etal-2020-findings-wmt}, the organizers adopt a version of span-level precision, recall, and $\fscore$ similar to \mpp.\footnote{The official WMT19 meta-evaluation script can be found here: \url{https://github.com/deep-spin/qe-evaluation/blob/master/eval_document_annotations.py}.} Each hypothesis span $\hyp \in \hyps$ is matched with the ground-truth error span with which it has the highest character overlap, and vice versa. The ``best match'' of each span is defined as follows: 
\begin{align*}
    bm(\hyp) &= \argmax_{s \in S} |\hyp \cap s| \\
    bm(s) &= \argmax_{\hyp \in \hyps} |\hyp \cap s|
\end{align*}
For each $\hyp \in \hyps$ and $s \in S$, span precision and recall are defined as follows:
\begin{align*}
    P_{\textsc{w19}}(\hyp) &= \frac{\left| \hyp \cap bm(\hyp) \right|}{\left| \hyp \right|} \\
    R_{\textsc{w19}}(s) &= \frac{\left| s \cap bm(s) \right|}{\left| s \right|}
\end{align*}
Document precision and recall are computed as the average span precision and recall:
\begin{align*}
    P_{\textsc{w19}} &= \frac{1}{|\hyps|} \sum_{\hyp \in \hyps} P_{\textsc{w19}}(\hyp) \\
    R_{\textsc{w19}} &= \frac{1}{|S|} \sum_{s \in S} R_{\textsc{w19}}(s)
\end{align*}
Finally, document $\fscore$ is the harmonic mean of average precision and recall, and corpus $\fscore$ is obtained by averaging $\fscore$ across documents.

Similar to \mpp, this measure assigns partial credit to partial overlaps. However, instead of computing a one-to-one matching between hypothesis and ground-truth error spans, each hypothesis span is matched with the ground-truth span with the largest character overlap. As a consequence, the same hypothesis span can be matched to multiple ground-truth error spans, and vice versa, and matches involving longer error spans are preferred over those involving shorter ones. Furthermore, corpus $\fscore$ is obtained via macro-averaging, which, in the presence of short documents and sparse human annotations, favors low-recall auto-evaluators, as shown in Section~\ref{sec:low-recall-results}.

\subsection{WMT 2023, 2024, and 2025} \label{apx:wmt25_measure}
The organizers of the 2023 and 2024 editions of the WMT Quality Estimation shared tasks \cite{blain-etal-2023-findings, zerva-etal-2024-findings} adopt a different version of span-level precision, recall, and $\fscore$, which was later slightly modified again for the 2025 edition of the WMT shared task on Automated Translation Quality Evaluation \cite{lavie-etal-2025-findings}. These measures are very similar to the one we dub $\approx$\textsc{w25} in Section~\ref{sec:mpp}, with the primary difference that they do not enforce a one-to-one matching between hypothesis and ground-truth error spans. We define them formally as follows.

Let us first define two functions to count the number of times each character index of translation $\bs{x} = (x_1, x_2, \ldots, x_n)$ participates in an error span:
\begin{align*}
    c_{\hyp}(i) &= \sum_{\hyp \in \hyps, \hyp = (i_{\hyp}, j_{\hyp})} \mathbb{I}[i_{\hyp} \leq i \leq j_{\hyp}] \\
    c_s(i) &= \sum_{s \in S, s = (i_s, j_s)} \mathbb{I}[i_s \leq i \leq j_s]
\end{align*}
We then define two binary functions that indicate whether a character index participates in \textit{any} error span:
\begin{align*}
    b_{\hyp}(i) &= \mathbb{I}[c_{\hyp}(i) > 0] \\
     b_{s}(i) &= \mathbb{I}[c_{s}(i) > 0]
\end{align*}
Precision and recall, as used by the WMT 2023 and 2024 Quality Estimation shared tasks, are defined as follows:\footnote{The official WMT23 meta-evaluation script can be found here: \url{https://github.com/WMT-QE-Task/qe-eval-scripts/blob/main/wmt23/task2_evaluate.py}.}
\begin{align*}
    P_{\textsc{w}23} = \frac{\sum_{i=1}^{n} b_{\hyp}(i) b_s(i)}{\sum_{i=1}^{n} b_{\hyp}(i)} \\
    R_{\textsc{w}23} = \frac{\sum_{i=1}^{n} b_{\hyp}(i) b_s(i)}{\sum_{i=1}^{n} b_s(i)}
\end{align*}

Later, at the WMT 2025 Automated Translation Quality Evaluation shared task, the organizers allowed multiple overlapping annotations to contribute as many times as each character participates in a distinct error span. Consequently, their version of precision and recall is defined directly using the counts $c(\cdot)$ rather than the binary functions $b(\cdot)$:\footnote{The official WMT25 meta-evaluation script can be found here: \url{https://github.com/wmt-conference/wmt25-mteval/blob/main/scripts/scoring/task2/scoring_esa.py}.}
\begin{align}
    P_{\textsc{w}25} = \frac{\sum_{i=1}^{n} \min(c_{\hyp}(i), c_s(i))}{\sum_{i=1}^{n} c_{\hyp}(i)} \\
    R_{\textsc{w}25} = \frac{\sum_{i=1}^{n} \min(c_{\hyp}(i), c_s(i))}{\sum_{i=1}^{n} c_s(i)}
\end{align}

Despite being used for span-level evaluation, both measures operate at the character level. As a result, longer error spans receive greater weight than shorter ones, making these measures unaligned with MT evaluation protocols such as ESA and MQM, where the weight of each error is determined by explicit severity labels rather than span length.

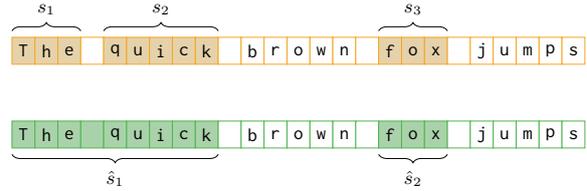
\begin{figure}
    \centering
    \resizebox{\linewidth}{!}{
    \begin{tikzpicture}[
        cell/.style={
          minimum width=0.38cm,
          minimum height=0.45cm,
          draw=#1,
          line width=0.3pt,
          outer sep=0pt,
          inner sep=0pt,
          font=\ttfamily\small,
          anchor=center
        },
        T empty/.style={cell=orangeSpan, fill=white},
        T filled/.style={cell=orangeSpan, fill=highlightT},
        S empty/.style={cell=greenSpan, fill=white},
        S filled/.style={cell=greenSpan, fill=highlightS},
        brace above/.style={
          decorate,
          decoration={brace, amplitude=4pt, raise=3pt}
        },
        brace below/.style={
          decorate,
          decoration={brace, amplitude=4pt, raise=3pt, mirror}
        },
        span label/.style={font=\footnotesize}
      ]
    
      \def\cellwidth{0.38}
      \def\rowsep{1.4}
      
      % =============================================
      % Character data: char/T-highlight/S-highlight
      % =============================================
      
      \foreach \char/\tstat/\sstat [count=\i from 0] in {
        T/1/1, h/1/1, e/1/1, { }/0/1,           % 0-3:   "The "
        q/1/1, u/1/1, i/1/1, c/1/1, k/1/1,      % 4-8:   "quick"
        { }/0/0,                                 % 9:     " "
        b/0/0, r/0/0, o/0/0, w/0/0, n/0/0,      % 10-14: "brown"
        { }/0/0,                                 % 15:    " "
        f/1/1, o/1/1, x/1/1,                    % 16-18: "fox"
        { }/0/0,                                 % 19:    " "
        j/0/0, u/0/0, m/0/0, p/0/0, s/0/0       % 20-24: "jumps"
        % { }/0/0,                                 % 25:    " "
        % o/0/0, v/0/0, e/0/0, r/0/0,             % 26-29: "over"
        % { }/0/0,                                 % 30:    " "
        % t/0/0, h/0/0, e/0/0,                    % 31-33: "the"
        % { }/0/0,                                 % 34:    " "
        % l/1/0, a/1/0, z/1/0, y/1/0,             % 35-38: "lazy"
        % { }/0/0,                                 % 39:    " "
        % d/0/1, o/0/1, g/0/1                     % 40-42: "dog"
      } {
        % T row (top)
        \ifnum\tstat=1
          \node[T filled] (T\i) at (\i*\cellwidth, 0) {\char};
        \else
          \node[T empty] (T\i) at (\i*\cellwidth, 0) {\char};
        \fi
        % S row (bottom)
        \ifnum\sstat=1
          \node[S filled] (S\i) at (\i*\cellwidth, -\rowsep) {\char};
        \else
          \node[S empty] (S\i) at (\i*\cellwidth, -\rowsep) {\char};
        \fi
      }
    
      % =============================================
      % T ANNOTATIONS (top braces)
      % Draw left to right so brace bulges upward
      % =============================================
      \draw[brace above] (T0.north west) -- (T2.north east)
        node[midway, above=7pt, span label] {$s_1$};
      \draw[brace above] (T4.north west) -- (T8.north east)
        node[midway, above=7pt, span label] {$s_2$};
      \draw[brace above] (T16.north west) -- (T18.north east)
        node[midway, above=7pt, span label] {$s_3$};
    
      % =============================================
      % S ANNOTATIONS (bottom braces)
      % Using mirror decoration so brace bulges downward
      % =============================================
      \draw[brace below] (S0.south west) -- (S8.south east)
        node[midway, below=7pt, span label] {$\hat{s}_1$};
      \draw[brace below] (S16.south west) -- (S18.south east)
        node[midway, below=7pt, span label] {$\hat{s}_2$};
    
    \end{tikzpicture}
    }
    \caption{Given the translation $\bs{x}=$ ``The quick brown fox jumps'', this example shows span-level annotations of translation quality, featuring three ground-truth error spans -- $s_1, s_2, s_3 \in S$ -- and two hypothesis error spans -- $\hyp_1, \hyp_2 \in \hyps$.}
    \label{fig:example-overlapping}
\end{figure}

\section{Effects of Enforcing a One-to-one Error Matching} \label{apx:enforcing-one-to-one}
In Section~\ref{sec:span-level-meta-evaluation}, we enforce a one-to-one matching between hypothesis and ground-truth error spans. In contrast, previous work has often implemented meta-evaluation measures without enforcing such a matching. For example, considering the measures employed over the years at WMT (Appendix~\ref{apx:wmt-measures}), \textsc{w19} computes two many-to-one alignments, different between precision and recall, while \textsc{w23} and \textsc{w25} do not align hypothesis and ground-truth errors at all, because the final precision and recall are directly based on character overlaps. Let us clarify the difference between these choices using the example in Figure~\ref{fig:example-overlapping}, where $\hyp_1$ overlaps with both $s_1$ and $s_2$. 

Let us start from \textsc{w19}, where each hypothesis span is aligned to the ground-truth span with the highest character overlap, and vice-versa (Section~\ref{apx:wmt19_and_20}). Consequently, $\hyp_1$ is paired to $s_2$ when computing precision, while both $s_1$ and $s_2$ are paired to $\hyp_1$ when computing recall, leading to the following $P$ and $R$ values:
\begin{align*}
    P_{\textsc{w19}} &= \frac{\frac{5}{9} + 1}{2} = \frac{7}{9} \\
    R_{\textsc{w19}} &= \frac{1 + 1 + 1}{3} = 1 
\end{align*}

In \textsc{w23} and \textsc{w25}, there is no span alignment, and $P$ and $R$ are based directly on character overlap counts. Thus, $\hyp_1$ overlaps with ground-truth spans for $8$ out of $9$ characters, while $s_1$ and $s_2$ overlap entirely with hypothesis error spans ($\hyp_1$ covers them both), leading to: 
\begin{align*}
    P_{\textsc{w25}} &= \frac{8 + 3}{9 + 3} = \frac{11}{12} \\
    R_{\textsc{w25}} &= \frac{3 + 5 + 3}{3 + 5 + 3} = 1 
\end{align*}

Instead, \mpp enforces a one-to-one error matching and selects the $M \in \mathcal{M}_{\mpp}$ that maximizes the final $\fscore$. In the example in Figure~\ref{fig:example-overlapping}, $\hyp_1$ is matched to $s_2$, while $s_1$ remains unmatched:
\begin{align*}
    P_{\mpp} &= \frac{\frac{5}{9} + 1}{2} = \frac{7}{9} \\
    R_{\mpp} &= \frac{1 + 1}{3} = \frac{2}{3}
\end{align*}

Arguably, one strategy is not necessarily better than the other, and the choice might depend on the specific evaluation objectives. We decided to enforce a one-to-one matching to align with the MQM protocol, enabling the maximum score of $F=1$ to be reached only when both the \textit{position and number} of detected errors match the ground truth. In contrast, measures that do not enforce a one-to-one matching focus primarily on error location. Consider again our running example in Figure~\ref{fig:example-overlapping}: auto-evaluators assessed using \textsc{w23} and \textsc{w25} can achieve an optimal $\fscore=1$ by returning as many single-character annotations as there are annotated characters in the ground-truth. However, the MQM protocol assigns scalar scores to translations based on both error severity and number. Specifically, a translation containing $11$ errors $e_i$ (i.e., the number of characters of ``The'',  ``quick'', and ``fox'') is assigned an MQM score of $-\sum_{i=1}^{11} \text{severity}(e_i)$, which is substantially worse than a translation containing three errors, even if their lengths and position match precisely.

\textsc{w19} mitigates this by aligning hypothesis and ground-truth errors based on their character overlap. However, it does not enforce a single one-to-one matching; instead, it computes two many-to-one alignments when computing precision and recall. This allows the same hypothesis error to be aligned to multiple (potentially distinct) ground-truth errors, and vice versa. Returning to our running example, this allows $R=1$ even though the number of hypothesis errors is smaller than the number of ground-truth errors. 

\section{Additional Results} \label{apx:additional-results}
Table~\ref{tab:wmt25-stats-mqm} shows the number of errors, average number of errors per sample, and average span length, for each evaluator, computed on the MQM split of the WMT 2025 data.

Table~\ref{tab:wmt25-mqm-scores} presents the full results -- in terms of auto-evaluator $F$-scores -- on the MQM split of the WMT 2025 data.

\begin{table}[t]
\centering
\begin{tabular}{lrrr}
\toprule
\textbf{Auto-Evaluator} & \#E & \#E/S & SL \\
\midrule
MQM                                          & 16,297 & 2.21 & 10.7 \\
AutoLQA41.sec                                &  2,116 & 0.29 & 53.1 \\
\haiku                                       & 11,439 & 1.55 & 10.7 \\
\sonnet                                      & 17,584 & 2.38 &  7.3 \\
AutoLQAESA.sec                               &  8,549 & 1.16 & 36.5 \\
AIP.pri                                      &  6,606 & 0.90 &  5.1 \\
AutoLQA.pri                                  & 10,070 & 1.36 & 35.5 \\
AIP.sec                                      &  7,098 & 0.96 &  5.0 \\
\gptoss                                      & 14,308 & 1.94 & 10.8 \\
GemSpanEval.pri                              & 26,363 & 3.57 & 10.5 \\
XCOMET-XXL.bas                               & 26,676 & 3.61 &  5.6 \\
XCOMET-XL.bas                                & 30,824 & 4.18 &  5.0 \\
\qwen                                        & 30,563 & 4.14 &  9.6 \\
GemSpanEval-QE.sec                           & 31,815 & 4.31 &  9.6 \\
\bottomrule
\end{tabular}
\caption{Number of errors (``\#E''), average number of errors per sample (``\#E/S''), and average span length (``SL''), for each evaluator, computed on the MQM split of the WMT 2025 data \cite{lavie-etal-2025-findings}.}
\label{tab:wmt25-stats-mqm}
\end{table}

\begin{table*}[t]
\centering
\begin{tabular}{l|rrrr|rrrr}
\toprule
& \multicolumn{4}{c|}{Micro} & \multicolumn{4}{c}{Macro} \\
\textbf{Auto-evaluator} & \exm & \mpa & \mpp & \textsc{w}25 & \exm & \mpa & \mpp & \textsc{w}25 \\
\midrule
\cellcolor{highlight}\qwen & 6.99 & 28.33 & 22.37 & 23.31 & 7.21 & 23.75 & 17.45 & 18.33 \\
GemSpanEval-QE.sec & 7.76 & 25.87 & 20.60 & 22.10 & 14.99 & 27.33 & 22.11 & 22.61 \\
\cellcolor{highlight}Sonnet 4.5 & 7.98 & 25.05 & 19.92 & 18.37 & 12.28 & 23.84 & 18.87 & 19.11 \\
GemSpanEval.pri & 7.11 & 25.29 & 19.84 & 20.61 & 13.95 & 25.88 & 20.65 & 20.95 \\
XCOMET-XXL & 2.68 & 25.53 & 18.33 & 20.89 & 7.81 & 24.73 & 17.43 & 19.48 \\
AIP.sec & 8.69 & 24.43 & 18.16 & 13.91 & 22.12 & 31.86 & 27.30 & 26.87 \\
\cellcolor{highlight}Haiku 4.5 & 6.29 & 22.45 & 17.69 & 17.52 & 14.14 & 24.77 & 20.31 & 20.55 \\
XCOMET-XL & 2.32 & 24.72 & 17.47 & 21.14 & 6.67 & 23.89 & 16.22 & 18.91 \\
\cellcolor{highlight}gpt-oss 120b & 6.62 & 22.06 & 17.45 & 16.35 & 10.99 & 21.99 & 17.21 & 17.42 \\
AIP.pri & 6.04 & 23.18 & 15.93 & 13.20 & 22.73 & 32.65 & 27.90 & 27.71 \\
AutoLQA.pri & 3.15 & 21.49 & 14.95 & 16.26 & 11.78 & 24.44 & 17.56 & 18.82 \\
AutoLQAESA.sec & 2.82 & 19.92 & 14.01 & 15.69 & 12.81 & 24.57 & 18.14 & 19.27 \\
AutoLQA41.sec & 1.72 & 11.35 & 9.30 & 15.15 & 27.16 & 34.16 & 30.36 & 31.40 \\
\bottomrule
\end{tabular}
\caption{$\fscore$ of auto-evaluators on the MQM split of the WMT 2025 Automated Translation Quality Evaluation Systems shared task under \exm, \mpa, \mpp, and \textsc{w}25, with both micro- and macro-averaging. We run the auto-evaluators highlighted in gray; the others are submissions to WMT.}
\label{tab:wmt25-mqm-scores}
\end{table*}

\subsection{WMT25 ESA Ranking}
Table~\ref{tab:wmt25-full-results-esa} presents the ranking of auto-evaluators when using the ESA split of WMT25 -- including all the translation directions associated with ESA in Table~\ref{tab:data-statistics} -- using \mpp with micro-averaging. Results are aggregated across language directions by averaging language pair-specific precision, recall, and $\fscore$.

\begin{table}[t]
\centering
\begin{tabular}{lrrr}
\toprule
\textbf{Metric} & \multicolumn{1}{c}{$P$} & \multicolumn{1}{c}{$R$} & \multicolumn{1}{c}{$F$1} \\
\midrule
AIP.sec & $17.00$ & $14.75$ & $15.17$ \\
AIP.pri & $16.06$ & $13.02$ & $13.52$ \\
AutoLQA.pri & $10.61$ & $14.59$ & $10.95$ \\
AutoLQAESA.sec & $11.45$ & $13.50$ & $10.77$ \\
XCOMET-XXL.bas & $6.80$ & $28.45$ & $10.55$ \\
GemSpanEval-QE.sec & $5.98$ & $33.99$ & $9.95$ \\
GemSpanEval.pri & $5.91$ & $33.51$ & $9.81$ \\
XCOMET-XL.bas & $6.22$ & $25.90$ & $9.60$ \\
AutoLQA41.sec & $11.68$ & $7.27$ & $8.30$ \\
\bottomrule
\end{tabular}
\caption{Full results on the ESA split of WMT25 under \mpp with micro-averaging.}
\label{tab:wmt25-full-results-esa}
\end{table}

\clearpage
\onecolumn

\newtcblisting{promptbox}{
  breakable,
  enhanced,
  listing only,
  listing engine=listings,
  colback=white,
  colframe=black!60,
  boxrule=0.5pt,
  width=\textwidth,
  left=2pt,right=2pt,top=2pt,bottom=2pt,
  listing options={
    basicstyle=\ttfamily\scriptsize,
    breaklines=true,
    breakatwhitespace=false,
    columns=fullflexible,
    keepspaces=true,
    showstringspaces=false,
  },
}

\begin{center}
\captionof{table}{Prompt used to instruct the LLMs to conduct the error detection task. It is designed to take as input source language, source text, target language, and target text, and return the annotation as a JSON object.}
\label{tab:error-detection-prompt}
\begin{promptbox}
You will be provided with a source paragraph and its translation. A paragraph may contain one or more sentences. Your task is to identify all translation errors, assigning a category, subcategory, and severity level to each error. 

### Task Guidelines

- To identify an error, you must mark its span of text in the translation. Only in two special cases must the error be located in the source paragraph rather than the translation. These two special cases depend on the error category you assign to the identified error (refer to error categories and subcategories below):
    1. **Omission errors** (category='Accuracy' and subcategory='Omission'): Mark the missing span of text in the source paragraph.  
    2. **Source errors** (category='Source error' and subcategory='Source error'): Mark the problematic span of text in the source paragraph. Source errors are problems in the source paragraph itself, not translation errors (e.g., grammatical errors in the source paragraph). When source errors occur, do not penalize the translation by marking a corresponding translation error unless the translation introduced additional problems. 
  Apart from these two special cases, all errors must be located in the translation.    

- When identifying errors, be as fine-grained as possible. For example, if two consecutive words are each mistranslated, record two separate errors. However, if multiple errors occur in a single stretch of text and cannot be separated, record only the most severe error (refer to the available error severities below).  

- We will later derive the position of the identified error spans in the source or translation paragraphs via string matching. Therefore, report the identified error spans verbatim, without modifying or altering them in any way. 

- If it is not possible to reliably identify distinct errors because the translation is too badly garbled or is unrelated to the source, mark a single 'Unintelligible' error that spans the entire paragraph. There can be at most one 'Unintelligible' error per translation, and it should span the entire paragraph. Do not identify other errors if the 'Unintelligible' category is used.

### Error typology

You must select error categories and subcategories from the following error typology:
```
**Accuracy**
    - Addition: Translation includes information not present in the source.
    - Omission: Translation is missing content from the source.
    - Mistranslation: Translation does not accurately represent the source.
    - Untranslated text: Source text has been left untranslated when it should have been translated (note: use common sense and consider target language conventions, as some text like certain titles or certain proper names are typically left untranslated).  
    
**Fluency**
    - Punctuation: Incorrect punctuation (for locale or style).
    - Spelling: Incorrect spelling or capitalization.
    - Grammar: Problems with grammar, other than orthography.
    - Register: Wrong grammatical register (e.g., inappropriately informal pronouns).
    - Inconsistency: Internal inconsistency (not related to terminology).
    - Character encoding: Characters are garbled due to incorrect encoding.

**Terminology**
    - Inappropriate for context: Terminology is non-standard or does not fit the context.
    - Inconsistent use: Terminology is used inconsistently.

**Style**
    - Awkward: Translation has stylistic problems.

**Locale convention**
    - Address format: Wrong format for addresses.
    - Currency format: Wrong format for currency.
    - Date format: Wrong format for dates.
    - Name format: Wrong format for names.
    - Telephone format: Wrong format for telephone numbers.
    - Time format: Wrong format for time expressions.

**Other**
    - Other: Any other issue.

**Source error**
    - Source error: An error in the source.

**Unintelligible**
    - Unintelligible: Impossible to reliably characterize distinct errors.
```
Each error category (e.g., Accuracy or Fluency) has one or more subcategories (such as Addition for Accuracy, and Punctuation for Fluency).

### Severity levels

You must select severity levels from the following list:
```
- **Critical**: Errors that severely distort the meaning of the source text or make the translation very difficult to understand or parse.
- **Major**: Errors that alter the meaning of the source or impact the readability or flow of the translation.
- **Minor**: Small imperfections that have minimal impact on meaning preservation or readability.
```

### Output Annotation Format
Return your annotations in JSON format as a list of Python dictionaries enclosed between triple backticks. Each dictionary represents a translation error and has the following form:
```json
{{
    "span": <minimal span of text containing the error>,
    "span_with_context": <extended span of text containing the error>,
    "explanation": <justification for marking this span as error>,
    "category": <error category>,
    "subcategory": <error subcategory>,
    "severity": <error severity>
}}
```
If no errors are found, return an empty list.

## Input Source and Translated Segments
The source paragraph and translation to evaluate are provided below:
```
{src_lang} source: {src}
{tgt_lang} translation: {tgt}
```

## Task instructions

You must execute these steps in order:
1. **ERROR IDENTIFICATION**: Analyze the translation sentence by sentence. For each sentence, quote it, then identify potential translation errors by specifying error spans within the considered sentence. List all the spans of text that are potential translation errors.
2. **ERROR ANALYSIS**: Examine each identified span in isolation:
    1. **REASONING**: Explain why this span of text should be considered an error. If during your reasoning you determine that this is not actually an error, discard it and move to the next span. Otherwise, proceed with the next step.
    2. **CATEGORIZATION**: Assign an appropriate category, a subcategory, and a severity level to the identified error. Pay particular attention to severity assignment. Ensure that the assigned severity label reflects the severity description. Adjust the severity level if your initial assessment doesn't align with the definitions. 
    3. **VERIFICATION**: Review the error you have identified, its category, subcategory, and severity level. Confirm compliance with the annotation guidelines by checking:
        - Was the error span correctly marked in the translation? Or is it an omission or source error and should be marked in the source?
        - Was the error span correctly copied verbatim from the translation or source paragraphs, or have other characters been added? 
3. **FINAL ANNOTATION GENERATION**: Generate the output annotation in JSON format as requested.

{final_instruction}
\end{promptbox}
\end{center}

\clearpage
\twocolumn

\end{document}